\documentclass[10pt,twocolumn,letterpaper]{article}

\usepackage{iccv}
\usepackage{times}
\usepackage{epsfig}
\usepackage{graphicx}
\usepackage{amsmath}
\usepackage{amssymb}
\usepackage{graphicx}  \graphicspath{{figures/}}
\usepackage{subcaption}
\usepackage{algorithm}
\usepackage{algpseudocode}
\usepackage[normalem]{ulem}
\usepackage{makecell}
\usepackage{rotating}
\usepackage{booktabs}
\usepackage{gensymb}
\usepackage{authblk}
\usepackage{capt-of}
\usepackage{enumitem}
\usepackage[dvipsnames,svgnames,x11names]{xcolor}
\usepackage[symbol]{footmisc}

\newcommand\blfootnote[1]{%
	\begingroup
	\renewcommand\thefootnote{}\footnote{#1}%
	\addtocounter{footnote}{-1}%
	\endgroup
}

\makeatletter
\renewcommand\AB@affilsepx{, \protect\Affilfont}
\makeatother

\usepackage[pagebackref=true,breaklinks=true,letterpaper=true,colorlinks,bookmarks=false]{hyperref}

 \iccvfinalcopy 

\newcommand{\setmode}[1]{\def\mode{#1}}
\setmode{final} 
\long\def\IGNORE#1{} \long\def\COMMENT#1{}
\def\authornote#1#2#3{{\textcolor{#2}{\textsl{\small#1:[*#3*]}}}}

\ifthenelse{\equal{\mode}{draft}} { 
    \newcommand{\jgnote}[1]{\authornote{JG}{Blue}{#1}} 
    \newcommand{\khnote}[1]{\authornote{KH}{Green}{#1}} 
    \newcommand{\glnote}[1]{\authornote{GL}{Orange}{#1}} 
    \newcommand{\jknote}[1]{\authornote{JK}{Brown}{#1}} 
        } {}

\ifthenelse{\equal{\mode}{final}} {
    \newcommand{\advise}[1]{}
    \newcommand{\clnote}[1]{}
    \newcommand{\jgnote}[1]{}
    \newcommand{\khnote}[1]{}
    \newcommand{\glnote}[1]{}
    \newcommand{\jknote}[1]{}
    \typeout{************* MODE: Final}
    } {}

\ificcvfinal\fi

\title{Neural Inverse Rendering of an Indoor Scene from a Single Image}

\author[1,2,3,\dag]{Soumyadip Sengupta}
\author[1,4,\dag]{Jinwei Gu}
\author[1]{Kihwan Kim}
\author[1]{Guilin Liu}
\author[2]{David W. Jacobs} 
\author[1]{Jan Kautz}

\affil[1]{NVIDIA}
\affil[2]{University of Maryland, College Park} 
\affil[3]{University of Washington}
\affil[4]{SenseTime}

\begin{document}
	
	\maketitle
	\thispagestyle{empty}
	
\begin{abstract}
	Inverse rendering aims to estimate physical attributes of a scene, \eg, reflectance, geometry, and lighting, from image(s). Inverse rendering has been studied primarily for single objects or with methods that solve for only one of the scene attributes. We propose the first learning based approach that jointly estimates albedo, normals, and lighting of an indoor scene from a single image. Our key contribution is the Residual Appearance Renderer (RAR), which can be trained to synthesize complex appearance effects (\eg, inter-reflection, cast shadows, near-field illumination, and realistic shading), which would be neglected otherwise. This enables us to perform self-supervised learning on real data using a reconstruction loss, based on re-synthesizing the input image from the estimated components. We finetune with real data after pretraining with synthetic data. To this end we use physically-based rendering to create a large-scale synthetic dataset, which is a significant improvement over prior datasets. Experimental results show that our approach outperforms state-of-the-art methods that estimate one or more scene attributes.
\end{abstract}

\section{Introduction}
\label{sec:intro}

\begin{figure}[t]
	\centering
	\includegraphics[width=0.45\textwidth]{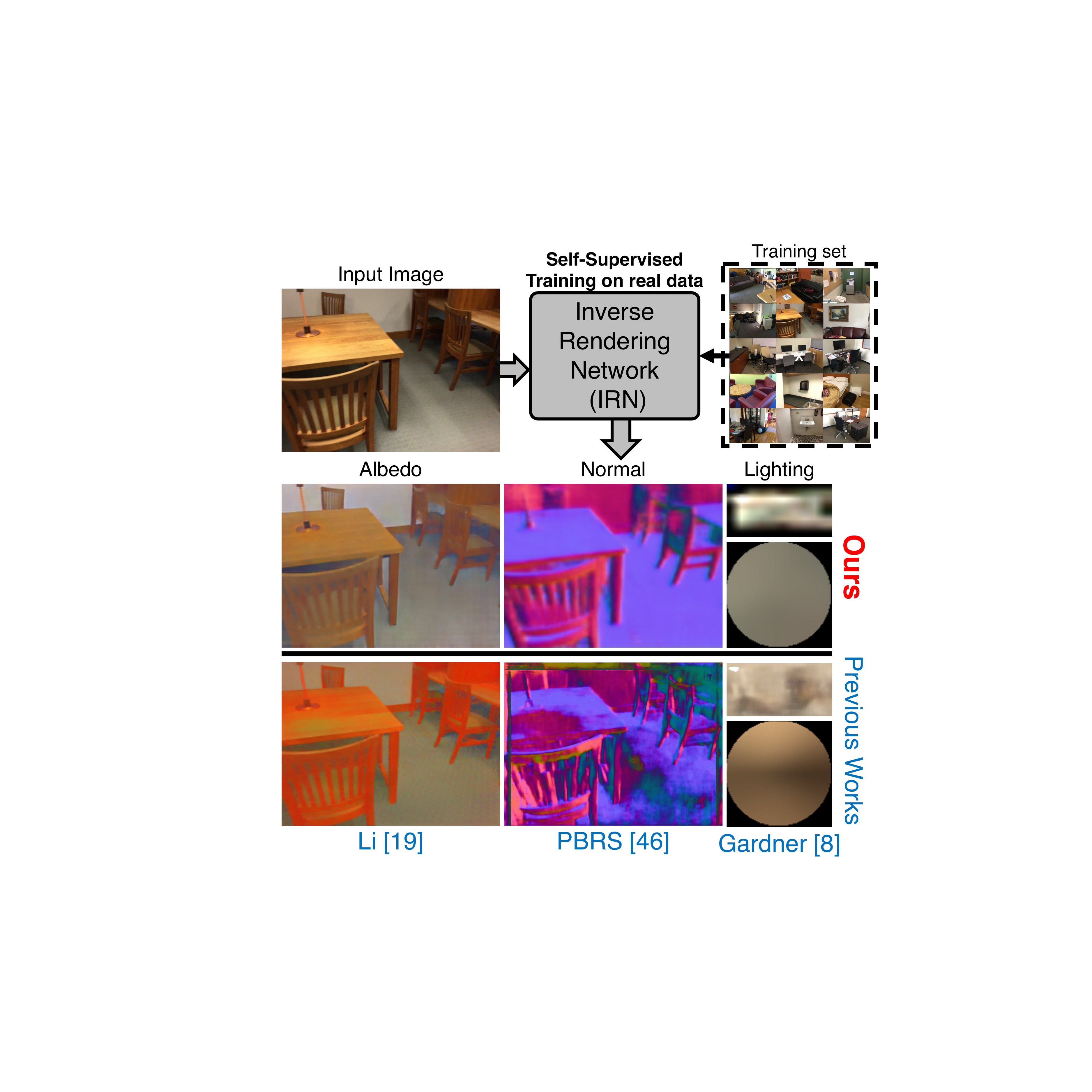}
	\caption{\small We propose a self-supervised approach for inverse rendering. We jointly decompose an indoor scene image into  albedo, surface normal and environment map lighting (top). Our method outperforms state-of-the-art approaches (bottom) that solve for only one of the scene attributes, \ie albedo (Li~\cite{li2018cgintrinsics}), normal (PBRS~\cite{zhang2016physically}) and lighting (Gardner~\cite{gardner2017learning}). }
	\label{fig:teaser}
	\vspace{-1em}
\end{figure}

Inverse rendering aims to estimate physical attributes (\eg, geometry, reflectance, and illumination) of a scene from images. It is one of the core problems in computer vision, with a wide range of applications in gaming, AR/VR, and robotics~\cite{Karsch:SA:11,khan2006image,tunwattanapong2009interactive,xu2018deep}. In this paper, we propose a holistic, data-driven approach for inverse rendering of an indoor scene from a single image. Inverse rendering has two main challenges. First, it is inherently ill-posed, especially if only a single image is given. Previous approaches~\cite{BarronTPAMI2015,kim2017lightweight,material2017,Meka:2018} that aim to solve this problem from a single image focus only on a single object. Second, inverse rendering of a \emph{scene} is particularly challenging, compared to single objects, due to complex appearance effects (\eg, inter-reflection, cast shadows, near-field illumination, and realistic shading). Some existing works~\cite{gardner2017learning,li2018materials,zhang2016physically,li2018cgintrinsics} are limited to estimating only one of the scene attributes. We focus on jointly estimating all scene attributes from a single image, which outperforms previous approaches that estimate a single attribute and generalizes better across datasets (see Figure \ref{fig:teaser} and more in Section~\ref{sec:result_comparison}).

A major challenge\blfootnote{\dag The authors contributed to this work when they were at NVIDIA.} in solving this problem is the lack of ground-truth labels for real images. Although we have ground-truth labels for geometry, collected by depth sensors, it is extremely difficult to measure reflectance and lighting at the large scale needed for training a neural network. Networks trained on synthetic images often fail to generalize well on real images. In this paper, we propose two key innovations aimed to tackle the domain gap between synthetic and real images. First, we propose the Residual Appearance Renderer (RAR), which learns to predict complex appearance effects on real images. This enables us to learn from unlabeled real images using self-supervised photometric reconstruction loss by re-synthesizing the image from its estimated components. Second, we introduce a new synthetic dataset using physically-based rendering with more photorealistic images than previous indoor scene datasets~\cite{zhang2016physically,song2016ssc}. More realistic synthetic data is useful for pretraining our network to help bridge the domain gap between real and synthetic images.

\noindent{\bf Residual Appearance Renderer}. Our key idea is to learn from unlabeled real data using self-supervised reconstruction loss, which is enabled by our proposed Residual Appearance Renderer (RAR) module. While using a reconstruction loss for domain transfer from synthetic to real has been explored previously~\cite{sfsnetSengupta18,shu2017neural,material2017}, their renderer is limited to direct illumination under distant lighting with a single material. For real images of a scene, however, this simple direct illumination renderer cannot synthesize important, complex appearance effects, such as inter-reflections, cast shadows, near-field lighting, and realistic shading. These effects termed \emph{residual appearance} in this paper, can only be simulated with the rendering equation via physically-based ray-tracing, which is non-differentiable and cannot be employed in a learning-based framework. To this end, we propose the Residual Appearance Renderer (RAR) module, which along with a direct illumination renderer can reconstruct the original image from the estimated scene attributes, for self-supervised learning on real images.

\noindent{\bf Rendering dataset}. It is especially challenging for inverse rendering tasks to obtain accurate ground truth labels for real images. Hence we create a large-scale synthetic dataset, \emph{CG-PBR} by applying physically-based rendering to all the 3D indoor scenes from SUNCG~\cite{song2016ssc}. Compared to prior work PBRS~\cite{zhang2016physically}, CG-PBR significantly improves data quality in the following ways: (1) The rendering of a scene is performed under multiple natural illuminations. (2) We render the same scene twice. Once with all materials set to Lambertian and once with the default material settings to produce image pairs (diffuse, specular). (3) We utilize deep denoising \cite{chaitanya2017interactive}, which allows us to render high-quality images from limited samples per pixel. Our dataset consists of 235,893 images with labels for normal, depth, albedo, Phong~\cite{lafortune1994using} model parameters and semantic segmentation. Examples are shown in Fig.~\ref{fig:data_show}. 
We plan to release the CG-PBR dataset upon acceptance of this paper.

Similar to prior works~\cite{li2018cgintrinsics,zhou2015learning}, we also make use of sparse labels on real data~\cite{bell2014intrinsic,Silberman:ECCV12}, whenever available, to further improve performance on real images.

To our knowledge, our approach is the first data-driven solution to single-image based inverse rendering of an indoor scene. SIRFS~\cite{BarronTPAMI2015}, which is a classical optimization based method, seems to be the only prior work with similar goals. Compared with SIRFS, as shown in Sec.~\ref{sec:result_comparison}, our method is more robust and accurate. In addition, we also compare with recent DL-based methods that estimate only one of the scene attributes, such as albedo~\cite{li2018cgintrinsics,li2018learning,nestmeyer2017reflectance,zhou2015learning}, lighting~\cite{gardner2017learning}, and normals~\cite{zhang2016physically}. Both quantitative and qualitative evaluations show that our approach outperforms these single-attribute methods and generalizes better across datasets, which seems to indicate that the self-supervised joint learning of all these scene attributes is helpful.

\section{Related Work}
\label{sec:related}

\paragraph{Optimization-based approaches.} For inverse rendering from a few images, most traditional optimization-based approaches make strong assumptions about statistical priors on illumination and/or reflectance. A variety of sub-problems have been studied, such as intrinsic image decomposition~\cite{tappen2003recovering}, shape from shading~\cite{prados2006shape,oxholm2012shape}, and BRDF estimation~\cite{lombardi2016reflectance}. Recently, SIRFS~\cite{BarronTPAMI2015} showed it is possible to factorize an image of an object or a scene into surface normals, albedo, and spherical harmonics lighting. In \cite{shelhamer2015scene} the authors use CNNs to predict the initial depth and then solve inverse rendering with an optimization. From an RGBD video, Zhang~\etal~\cite{zhang2016emptying} proposed an optimization method to obtain reflectance and illumination of an indoor room. These optimization-based methods, although physically grounded, often do not generalize well to real images where those statistical priors are no longer valid. 

\begin{figure*}[!h]
	\centering
	\includegraphics[width=0.99\textwidth]{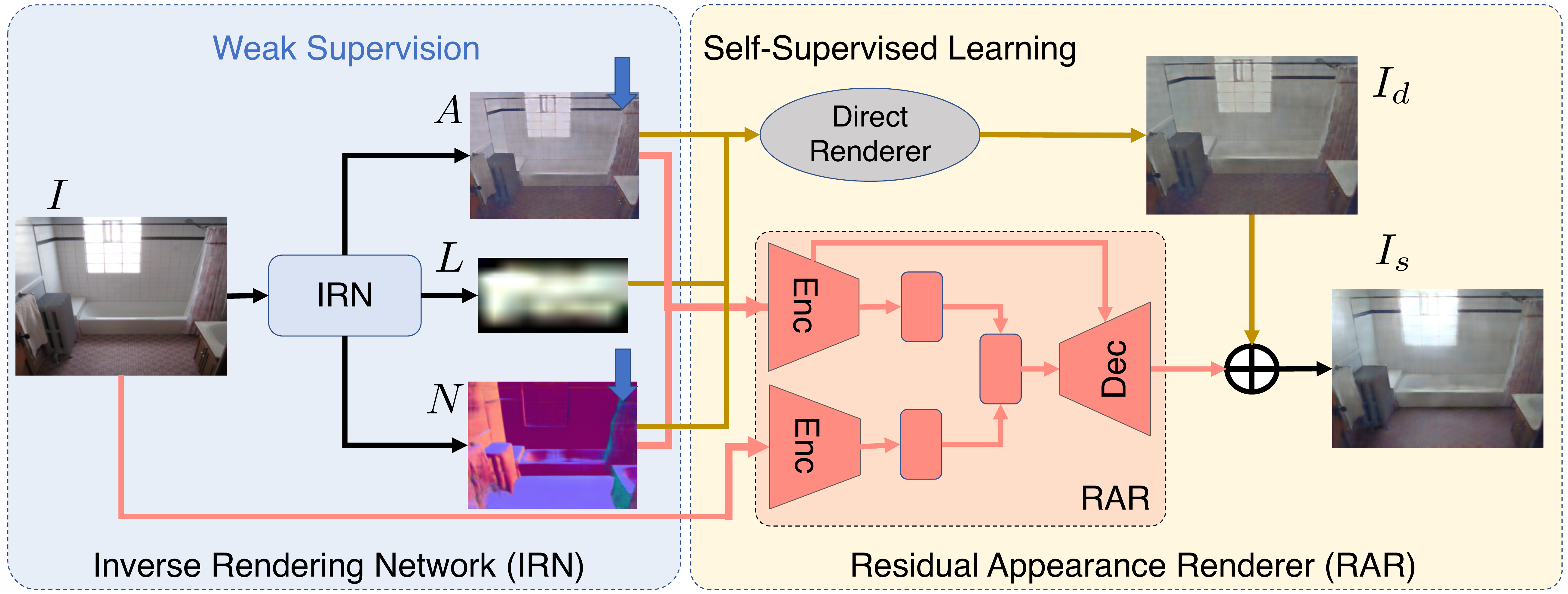}  
	\caption{\small \textbf{Overview of our approach.} Our Inverse Rendering Network (IRN) predicts albedo, normals and illumination map. We train on unlabeled real images using self-supervised reconstruction loss. Reconstruction loss consists of a closed-form Direct Renderer with no learnable parameters and the proposed Residual Appearance Renderer (RAR), which learns to predict complex appearance effects.} 
	\label{fig:networks}
\end{figure*}

\paragraph{Learning-based approaches.} With recent advances in deep learning, researchers have proposed to learn data-driven priors to solve some of these inverse problems with CNNs, many of which have achieved promising results. For example, it is shown that depth and normals may be estimated from a single image~\cite{eigen2015predicting, Fu2018DeepOR, zhuo2015indoor} or multiple images~\cite{Taniai2018}. Parametric BRDF may be estimated either from an RGBD sequence of an object ~\cite{Meka:2018,kim2017lightweight} or for planar surfaces~\cite{li2018materials}. Lighting may also be estimated from images, either as an environment map~\cite{gardner2017learning,hold2017deep}, or spherical harmonics~\cite{zhou2017label} or point lights~\cite{zhang2018discovering}. Recent works~\cite{shu2017neural,sfsnetSengupta18} also performed inverse rendering on faces. Some recent works also jointly learn some of the intrinsic components of an object, like reflectance and illumination~\cite{georgoulis2016delight,wang2018joint}, reflectance and shape~\cite{li2018svreflectance}, and normal, BRDF, and distant lighting ~\cite{shi2017learning,material2017}. Nevertheless, these efforts are mainly limited to objects rather than scenes, and do not model the aforementioned residual appearance effects such as inter-reflection, near-field lighting, and cast shadows present in real images. Concurrently, Yu \textit{et. al.} \cite{yu2019inverserendernet} proposed an inverse rendering framework for outdoor scenes using multi-view stereo as supervision.

\paragraph{Differentiable Renderer.} A few recent works from the graphics 
community proposed differentiable Monte Carlo renderers~\cite{Li:2018:DMC, che2018inverse} for optimizing rendering parameters (\eg, camera poses, scattering parameters) for synthetic 3D scenes. Neural mesh renderer~\cite{kato2018neural} addressed the problem of differentiable visibility and rasterization. Our proposed RAR is in the same spirit, but its goal is to synthesize the complex appearance effects for inverse rendering on \emph{real images}, which is significantly more challenging.

\paragraph{Datasets for inverse rendering.} High-quality synthetic data is essential for learning-based inverse rendering. SUNCG~\cite{song2016ssc} created a large-scale 3D indoor scene dataset. The images of the SUNCG dataset are not photo-realistic as they are rendered with OpenGL using diffuse materials and point source lighting. An extension of this dataset, PBRS~\cite{zhang2016physically}, uses physically based rendering to generate photo-realistic images. However, due to the computational bottleneck in ray-tracing, the rendered images are quite noisy and limited to one lighting condition. There also exist a few real-world datasets with partial labels on geometry, reflectance, or lighting. NYUv2~\cite{Silberman:ECCV12} provides surface normals from indoor scenes. Relative reflectance judgments from humans are provided in the IIW dataset~\cite{bell2014intrinsic} which are used in many intrinsic image decomposition methods. In contrast to these works, we created a large-scale synthetic dataset with significant image quality improvement.

\paragraph{Intrinsic image decomposition.} 

Intrinsic image decomposition is a sub-problem of inverse rendering, where a single image is decomposed into albedo and shading. Recent methods learn intrinsic image decomposition from labeled synthetic data~\cite{lettry2018darn,narihira2015direct,shi2017learning} and from unlabeled~\cite{li2018learning} or partially labeled real data~\cite{zhou2015learning,li2018cgintrinsics,nestmeyer2017reflectance,bell2014intrinsic}. Intrinsic image decomposition methods do not explicitly recover geometry or illumination but rather combine them together as shading. In contrast, our goal is to perform a complete inverse rendering which has a wider range of applications in AR/VR.

\section{Our Approach}
\label{sec:approach}

We present a deep learning-based approach for inverse rendering from a single image, which is shown in Fig.~\ref{fig:networks}. Given an input image $I$, our proposed neural Inverse Rendering Network (IRN), denoted as $h_d(\cdot;\Theta_d)$, estimates surface normal $N$, albedo $A$, and environment map $L$:
\begin{align}
\mbox{IRN}: &  \hspace{1em} h_d(I;\Theta_d)\rightarrow \left\{\hat{A},\hat{N},\hat{L}\right\}.
\vspace{-0.5em}
\end{align}

Using our synthetic data CG-PBR, we can simply train IRN ($h_d(\cdot;\Theta_d)$ with supervised learning -- with only one caveat, \ie we need to approximate the ``ground truth'' environment maps (using a separate network $h_e(\cdot;\Theta_e)\rightarrow \hat{L}^*$; see Sec.~\ref{subsec:approach_synthetic} for details). To generalize on real images, we use a self-supervised reconstruction loss. Specifically, as shown in Fig.~\ref{fig:networks}, we use two renderers to re-synthesize the input image from the estimations. The direct renderer $f_d(\cdot)$ is a simple closed-form shading function with no learnable parameters, which synthesizes the direct illumination part $\hat{I}_d$ of the the raytraced image. The Residual Appearance Renderer (RAR), denoted by $f_r(\cdot;\Theta_r)$, is a trainable network module, which learns to synthesize the complex appearance effects $\hat{I_r}$:
\begin{align}
\mbox{Direct Renderer}: & \hspace{1em} f_d(\hat{A},\hat{N},\hat{L}) \rightarrow \hat{I_d} \\
\mbox{RAR}: & \hspace{1em} f_r(I,\hat{A},\hat{N};\Theta_r) \rightarrow \hat{I_r}.
\end{align}
The self-supervised reconstruction loss is thus defined as $||I - (\hat{I_d} + \hat{I_r})||_1$. We explain the details of the direct renderer and the RAR in Sec.~\ref{subsec:approach_rar}.

\subsection{Training on Synthetic Data}
\label{subsec:approach_synthetic}

We first train IRN on our synthetic dataset CG-PBR with supervised learning. As shown in Fig.~\ref{fig:networks},  IRN has a structure similar to~\cite{sfsnetSengupta18}, which consists of a convolutional encoder, followed by nine residual blocks and a convolutional decoder for estimating albedo and normals. We condition the lighting estimation block on the image, normals and albedo features (see Appendix for details). We use ground truth albedos $A^*$ and normals $N^*$ from CG-PBR for supervised learning.

The ground truth environmental lighting $L^*$ is challenging to obtain, as it is the approximation of the actual surface light field.  We use environment maps as the exterior lighting for rendering CG-PBR, but these environment maps cannot be directly set as $L^*$, because the virtual cameras are placed \emph{inside} each of the indoor scenes. Due to occlusions, only a small fraction of the exterior lighting (\eg, through windows and open doors) is directly visible. The surface light field of each scene is mainly due to global illumination (\ie, inter-reflection) and some interior lighting. One could approximate $L^*$ by minimizing the difference between the raytraced image $I$ and the output $I_d$ of the direct renderer $f_d(\cdot)$ with ground truth albedo $A^*$ and normal $N^*$. However, we found this approximation to be inaccurate, since $f_d(\cdot)$ cannot model the residual appearance present in the raytraced image $I$. 

We thus resort to a learning-based method to approximate the ground truth lighting $L^*$. Specifically, we train a residual block based network, $h_e(\cdot;\Theta_e)$, to predict $\hat{L}^*$ from the input image $I$, normals $N^*$ and albedo $A^*$. We first train $h_e(\cdot;\Theta_e')$ with the images synthesized by the direct renderer $f_d(\cdot)$ with ground truth normals, albedo and indoor lighting, $I_d=f_d(A^*, N^*, L)$, where $L$ is randomly sampled from a set of real \emph{indoor} environment maps. Here the network learns a prior over the distribution of indoor lighting, \ie, $h(I_d;\Theta_e')\rightarrow L$. Next, we fine-tune this network $h_e(\cdot;\Theta_e')$ on the raytraced images $I$, by minimizing the reconstruction loss: $\|I - f_d(A^*, N^*, \hat{L}^*) \|$. Thus we obtain the approximated ground truth of the environmental lighting $\hat{L}^*=h_e(I;\Theta_e)$ which can best reconstruct the raytraced image $I$ modelled by the direct render.

Finally, with all the ground truth components ready, the supervised loss for training IRN is
\begin{equation}
\begin{split}
L_s &=  \lambda_1 ||\hat{N}-N^*||_1 + \lambda_2 ||\hat{A}-A^*||_1 \\
&+ \lambda_3 ||f_d(A^*,N^*, \hat{L})-f_d(A^*,N^*,\hat{L}^*)||_1.
\end{split}
\end{equation}
where $\lambda_1=1$, $\lambda_2=1$, and $\lambda_3=0.5$.

\subsection{RAR: Self-supervised Training on Real Images}
\label{subsec:approach_rar}

Learning from synthetic data alone is not sufficient to perform well on real images. Although CG-PBR was created with physically-based rendering, the variation of objects, materials, and illumination is still limited compared to those in real images. Since obtaining ground truth labels for inverse rendering is almost impossible for real images (especially for reflectance and illumination), we use two key ideas for domain transfer from synthetic to real: (1) self-supervised reconstruction loss and  (2) weak supervision from sparse labels.


\begin{figure*}[!h]
	\centering
	\includegraphics[width=0.98\textwidth]{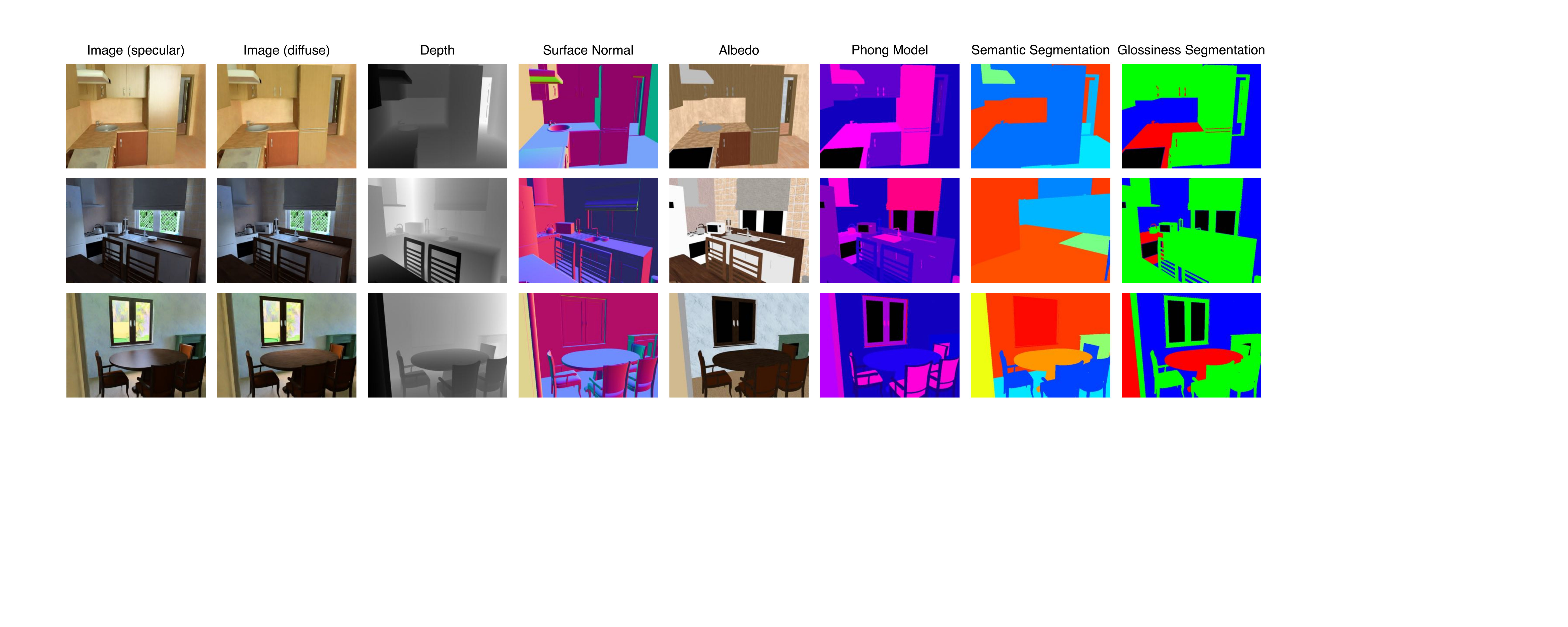} 
	\caption{\small \textbf{Our CG-PBR Dataset.} We provide 235,893 images of a scene assuming specular and diffuse reflectance along with ground truth depth, surface normals, albedo, Phong model parameters, semantic segmentation and glossiness segmentation.}
	\label{fig:data_show}	
\end{figure*}


Previous works on faces~\cite{sfsnetSengupta18,shu2017neural} and objects~\cite{material2017} have shown success in using a self-supervised reconstruction loss for learning from unlabeled real images. As mentioned earlier, the reconstruction in these prior works is limited to the direct renderer $f_d(\cdot)$, which is a simple closed-form shading function (under distant lighting) with no learnable parameters. In this paper, we implement $f_d(\cdot)$ simply as
\begin{equation}
\label{eq:direct_renderer}
\hat{I}_d = f_d(\hat{A},\hat{N},\hat{L}) = \hat{A}\sum_i \max(0,\hat{N}\cdot\hat{L}_i),
\vspace{-1em}
\end{equation}
where $\hat{L}_i$ corresponds to the pixels on the environment map $\hat{L}$. While using $f_d(\cdot)$ to compute the reconstruction loss may work well for faces~\cite{sfsnetSengupta18} or small objects with homogeneous material~\cite{material2017}, we found that it fails for inverse rendering of a scene. In order to synthesize the aforementioned residual appearances (\eg, inter-reflection, cast shadows, near-field lighting), we propose to use the differentiable Residual Appearance Renderer (RAR), $f_r(\cdot;\Theta_r)$, which learns to predict a residual image $\hat{I}_r$. The self-supervised reconstruction loss is thus defined as $L_u = ||I-(\hat{I}_d+\hat{I}_r)||_1$.

\begin{figure}[h]
	\centering
	\includegraphics[width=0.48\textwidth]{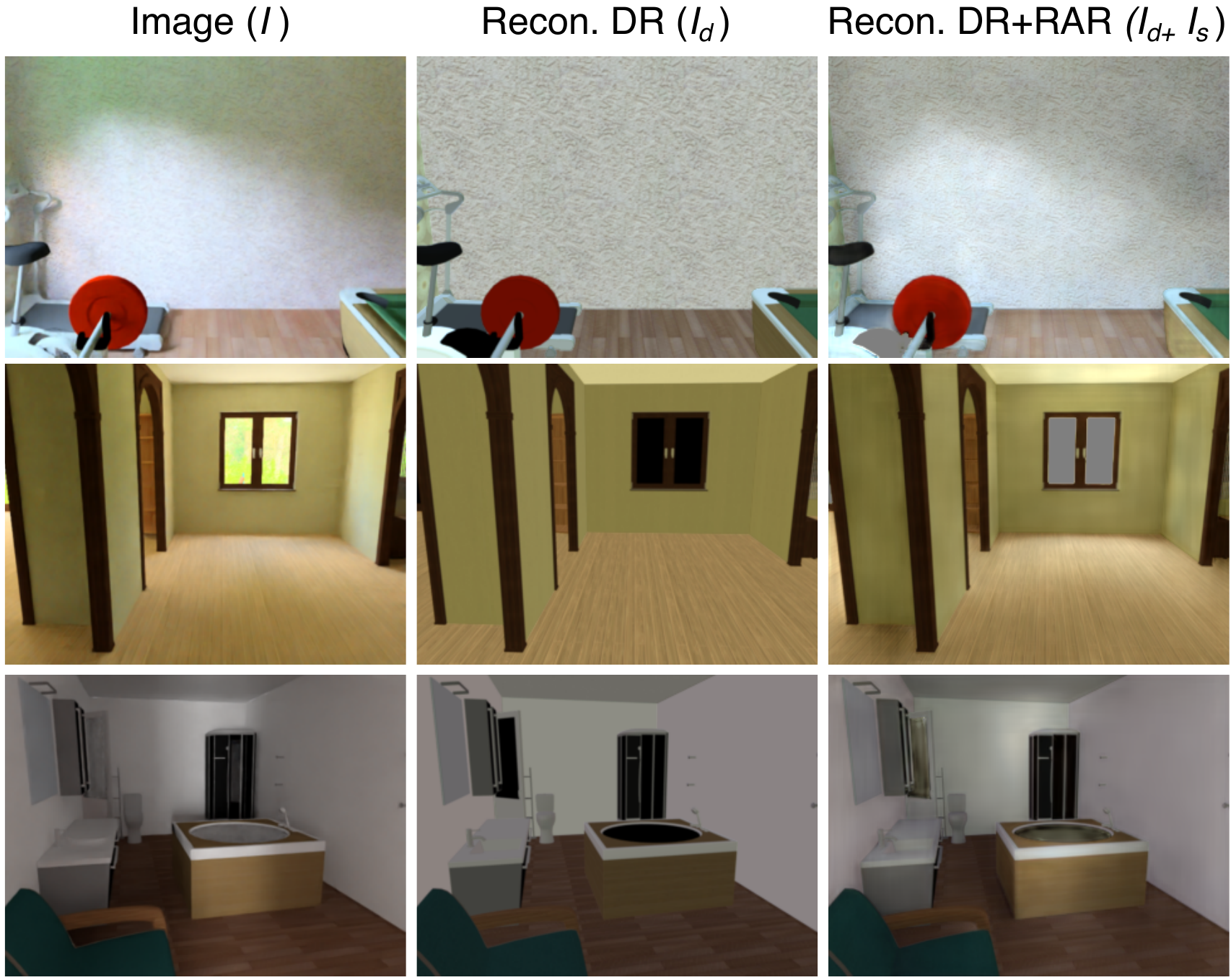}
	\caption{\small RAR $f_r(\cdot)$ learns to predict complex appearance effects  (\eg near-field lighting, cast shadows, inter-reflections) which cannot be modeled by a direct renderer (DR) $f_d(\cdot)$.}
	\label{fig:rar_learns}
\end{figure}

Our goal is to train RAR to capture only the residual appearances but \emph{not} to correct the artifacts of the direct rendered image due to faulty normals, albedo, and lighting estimation of the IRN. To achieve this goal, we train RAR \emph{only} on synthetic data with ground-truth normals and albedo, and fix it for training on real data, so that it only learns to correctly predict the residual appearances when the direct renderer reconstruction is accurate. We need realistic synthetic images, which capture complex appearance effects, \eg, cast shadows, inter-reflections etc.\ for training RAR. Our CG-PBR provides the necessary photo-realism that previous indoor scene datasets~\cite{zhang2016physically,song2016ssc} lack.

As shown in Fig.~\ref{fig:networks}, RAR consists of a U-Net~\cite{ronneberger2015u}, with normals and albedo as its input, and latent image features ($D=300$ dimension) learned by a convolutional encoder (`Enc'). We combine the image features at the end of the U-Net encoder and process them with the U-Net decoder to produce the residual image. RAR, along with the direct renderer $f_d(\cdot)$, acts like an auto-encoder in principle. RAR learns to encode complex appearance features from the original image into a latent subspace ($D = 300$ dimension). The bottleneck of the auto-encoder architecture present in RAR forces it to focus only on the complex appearance features and not in the whole image. So RAR learns to encode the non-direct part of the image to avoid paying a penalty in the reconstruction loss and in principle is simpler than a differentiable renderer. Details of the RAR architecture are presented in the Appendix (Section \ref{sec:appen}).

As shown in Fig.~\ref{fig:rar_learns}, RAR indeed learns to synthesize complex residual appearance effects present in the original input image. In Sec.~\ref{sec:result_ablation}, we provide quantitative and qualitative ablation studies to show why RAR is crucial in improving albedo and normal estimation. The goal of RAR in this project is to reconstruct an image from its components along with the direct renderer to facilitate self-supervised learning on real images. This helps to significantly improve albedo and normal estimation over state-of-the-art approaches. Our goal is \emph{not} to develop a differentiable renderer for realistic illumination during object insertion.

Similar to prior work~\cite{zhou2015learning,li2018cgintrinsics}, we use sparse labels over reflectance as weak supervision during training on real images. Specifically, we use pair-wise relative reflectance judgments from the Intrinsic Image in the Wild (IIW) dataset \cite{bell2014intrinsic} as a form of supervision over albedo. We also use supervision over surface normals from NYUv2 dataset~\cite{Silberman:ECCV12}. More details are provided in the Appendix (Section \ref{sec:appen}). As shown in Sec.~\ref{sec:result_ablation}, using such weak supervision can substantially improve performance on real images.

\subsection{Training Procedure}
\label{subsec:approach_protocol}

We summarize the different stages of training from synthetic to real data. More details are in the Appendix (Section \ref{sec:appen}).

\textbf{Estimate GT indoor lighting:} (a) First train $h_e(\cdot;\Theta_e')$ on images rendered by the direct renderer $f_d(\cdot)$. (b) Fine-tune $h_e(\cdot;\Theta_e)$ on raytraced synthetic images to estimate GT indoor environment map $\hat{L}^*$.

\textbf{Train on synthetic images:} (a) Train IRN with supervised L1 loss on albedo, normal and lighting. (b) Train RAR on synthetic data with L1 image reconstruction loss.

\textbf{Train on real images:} Fine-tune IRN on real data with (1) the pseudo-supervision over albedo, normal and lighting (to handle ambiguity of decomposition as proposed in~\cite{sfsnetSengupta18}), (2) the self-supervised reconstruction loss $L_u$ with RAR, and (3) the weak supervision over the albedo (\ie, pair-wise relative reflectance judgment) or normals.

\section{The CG-PBR Dataset}
\label{sec:dataset}

High-quality synthetic datasets are essential for learning-based inverse rendering. The SUNCG dataset~\cite{song2016ssc} contains 45,622 indoor scenes with 2,644 unique objects, but their images are rendered with OpenGL under fixed point light sources. The PBRS dataset~\cite{zhang2016physically} extends the SUNCG dataset by using physically based rendering with Mitsuba~\cite{Mitsuba}. Yet, due to a limited computational budget, many rendered images in PBRS are quite noisy. 
Moreover, the images in PBRS are rendered with only diffuse materials and a single outdoor environment map, which also significantly limits the photo-realism of the rendered images. High-quality photo-realistic images are necessary for training RAR to capture residual appearances.

\begin{figure}[!ht]
	\centering
	\includegraphics[width=0.48\textwidth]{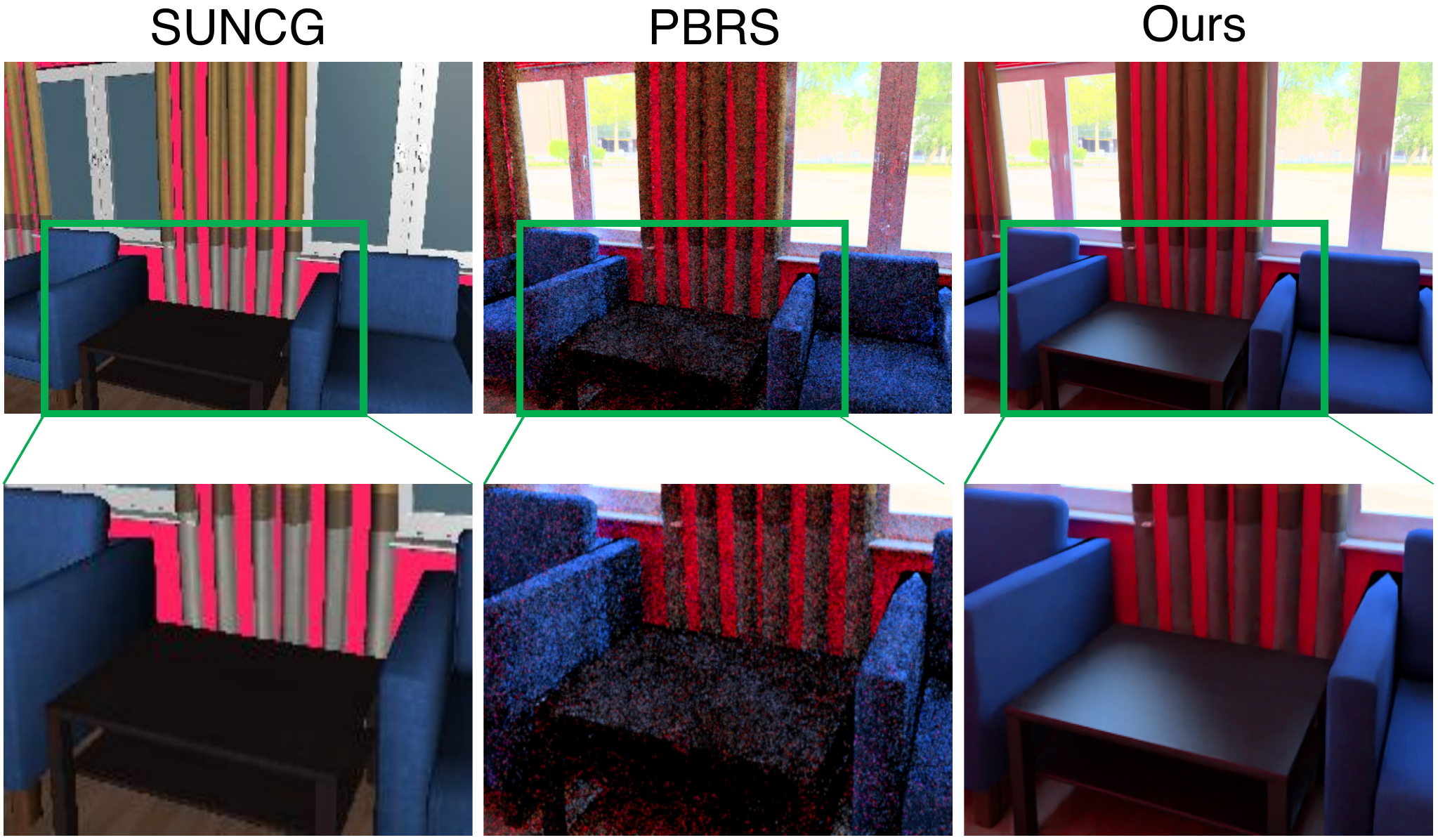}
	\caption{\small \textbf{Comparison with PBRS~\cite{zhang2016physically} and SUNCG~\cite{song2016ssc}.} Our dataset introduces more photo-realistic and less noisy images with specular highlights under multiple lighting conditions.}
	\label{fig:data_comp}	
\end{figure}

In this paper, we introduce a new dataset named CG-PBR, which improves data quality in the following ways: (1) The rendering is performed under multiple outdoor environment maps. (2) We render the same scene twice, once with all materials set to Lambertian and once with the default material settings. This offers (diffuse, specular) image pairs which can be useful to the community for learning to remove highlights and many other potential applications. (3) We utilize deep denoising \cite{chaitanya2017interactive}, which allows us to raytrace high-quality images from limited samples per pixel. Our dataset consists of 235,893 images with labels related to normal, depth, albedo, Phong~\cite{lafortune1994using} model parameters, semantic and glossiness segmentation. Examples are shown in Fig.~\ref{fig:data_show}.  A comparison with the SUNCG and PBRS datasets is shown in Fig.~\ref{fig:data_comp}. 

\section{Experimental Results}
\label{sec:result_comparison}

\paragraph{Comparison with SIRFS.} 

\begin{figure}[!h]
	\centering
	\includegraphics[width=\linewidth]{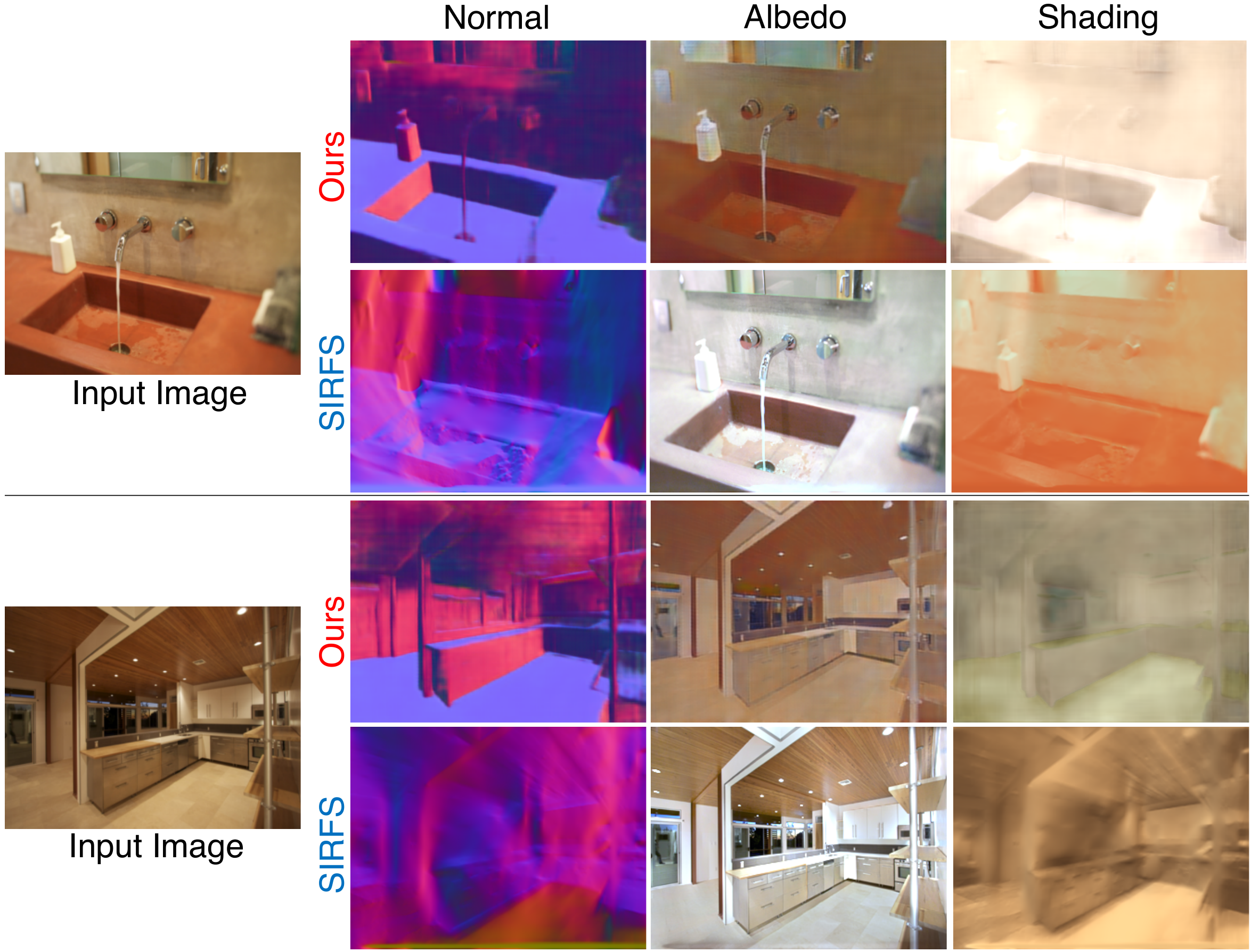}
	\caption{\small \textbf{Comparison with SIRFS~\cite{BarronTPAMI2015}.} Using deep CNNs our method performs better disambiguation of reflectance from shading and predicts better surface normals.}
	\label{fig:sirfs}
\end{figure}

SIRFS~\cite{BarronTPAMI2015} is an optimization-based method for inverse rendering, which estimates surface normals, albedo and spherical harmonics lighting from a single image. We compare with SIRFS on the test data from the IIW dataset~\cite{bell2014intrinsic}. As shown in Fig.~\ref{fig:sirfs}, our method produces more accurate normals and better disambiguation of reflectance from shading. This is expected, as we are using deep CNNs, which are known to better learn and utilize statistical priors present in the data than traditional optimization techniques.

\paragraph{Comparison with intrinsic image decomposition algorithms.} 

Intrinsic image decomposition aims to decompose an image into albedo and shading, which is a sub-problem in inverse rendering. Several recent works~\cite{bell2014intrinsic,zhou2015learning,nestmeyer2017reflectance,li2018cgintrinsics} showed promising results with deep learning. While our goal is to solve the complete inverse rendering problem, we still compare albedo prediction with these latest intrinsic image decomposition methods. We evaluate the WHDR (Weighted Human Disagreement Rate) metric~\cite{bell2014intrinsic} on the test set of the IIW dataset~\cite{bell2014intrinsic} and report the result in Table \ref{tab:iiw}. As shown, we outperform these algorithms that train on the original IIW dataset~\cite{bell2014intrinsic}. Since our goal is not intrinsic image decomposition, we do not train on additional intrinsic image specific datasets and avoid any post-processing as done in Li~\etal~\cite{li2018cgintrinsics}.

\begin{table}
	\centering
	\caption{\small \textbf{Intrinsic image decomposition on the IIW test set~\cite{bell2014intrinsic}}}. 
	\vspace{-1em}
	\begin{tabular}{ccc}
		\toprule
		Algorithm                           & Training set           & WHDR   \\ 
		\midrule
		Bell \textit{et. al.}~\cite{bell2014intrinsic} & - & 20.6\% \\ 
		Li \textit{et. al.}~\cite{li2018learning} & - & 20.3\% \\ 
		Zhou \textit{et. al.}~\cite{zhou2015learning} & IIW & 19.9\% \\ 
		Nestmeyer \textit{et. al.}~\cite{nestmeyer2017reflectance} & IIW & 19.5\% \\ 
		Li \textit{et. al.}~\cite{li2018cgintrinsics} & IIW & 17.5\% \\ 
		\textbf{Ours}                           & IIW & \textbf{16.7\%} \\ 
		\bottomrule
	\end{tabular}
	\label{tab:iiw}
\end{table}

\begin{table}[!h]
	\centering
	\caption{\small \textbf{Albedo estimation on synthetic data.} (RMSE;MAD)}
	\vspace{-.5em}
	\begin{tabular}{ccc}
		\toprule
		Algorithm  & CG-PBR        & CGI~\cite{li2018cgintrinsics}   \\ 
		\midrule
		Li \textit{et. al.}~\cite{li2018cgintrinsics} & 0.2873; 0.2427 & 0.2685; 0.2220 \\
		\textbf{Ours}                           & \textbf{0.1567; 0.1300} & \textbf{0.2126; 0.1875} \\ 
		\bottomrule
	\end{tabular}
	\label{tab:albedo}
\end{table}

\begin{figure}
	\centering
	\includegraphics[width=0.48\textwidth]{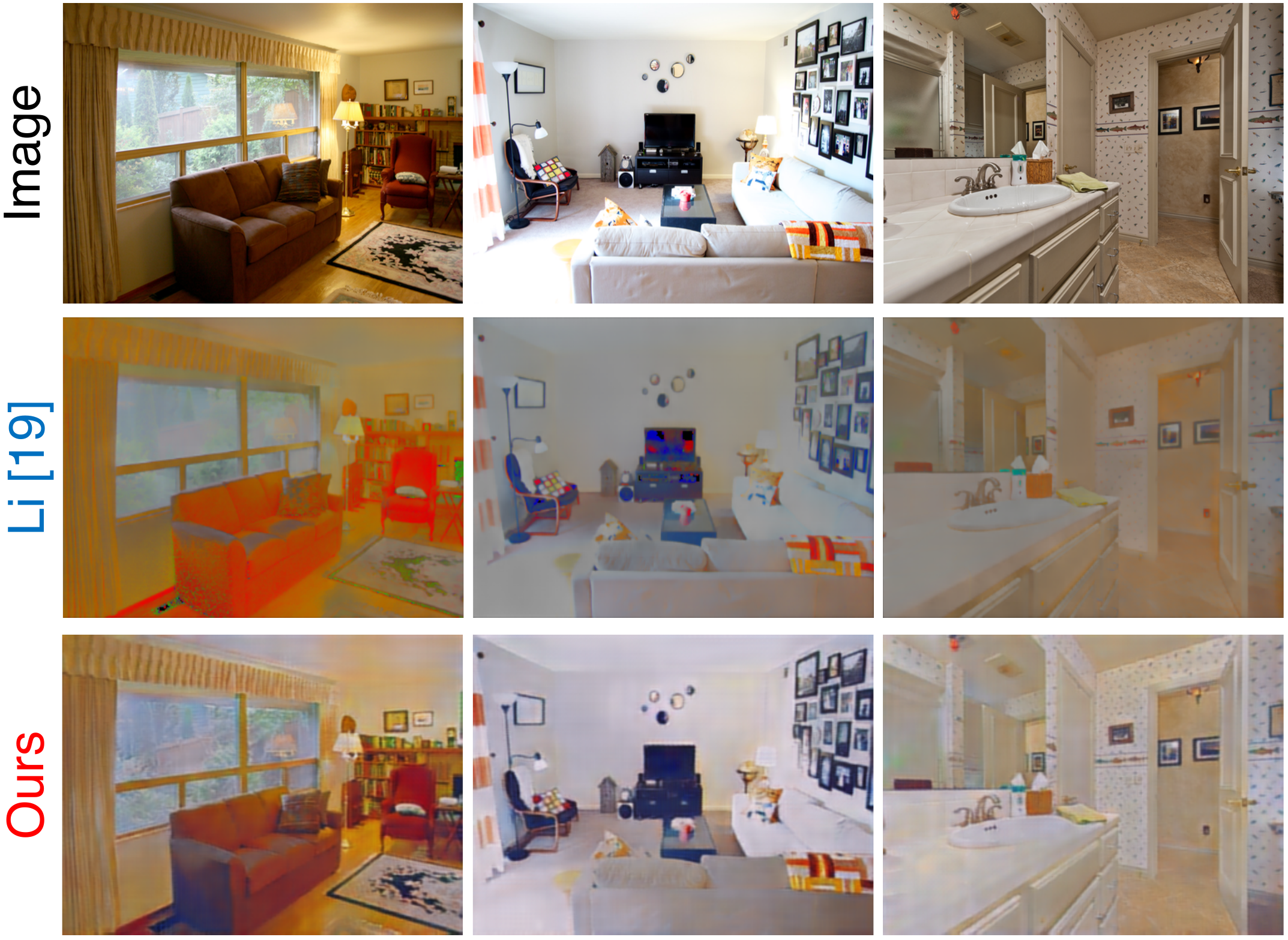}
	\caption{\small \textbf{Comparison with CGI (Li \textit{et. al.}~\cite{li2018cgintrinsics}). } In comparison with CGI~\cite{li2018cgintrinsics}, our method performs better disambiguation of reflectance from shading and preserves the texture in the albedo.}
	\label{fig:iiw_cgi}
\end{figure}

We also present a qualitative comparison of the inferred albedo with Li \etal~\cite{li2018cgintrinsics} in Figure~\ref{fig:iiw_cgi} (more in Appendix, Sec. \ref{sec:app_res}). As shown, our method preserves more detailed texture and has fewer artifacts in the predicted albedo, compared to  Li \etal~\cite{li2018cgintrinsics}. To further illustrate the effectiveness of our method over Li \etal~\cite{li2018cgintrinsics}, we also evaluate albedo estimation error (RMSE and MAD) on the synthetic CG-PBR and CGI datasets~\cite{li2018cgintrinsics} in Table \ref{tab:albedo} and qualitatively in Appendix, Sec. \ref{sec:app_res}.While our method uses CG-PBR as synthetic data for training, Li \etal~\cite{li2018cgintrinsics} uses CGI. Yet we outperform Li~\etal~\cite{li2018cgintrinsics} on both datasets. Thus our method significantly outperforms all prior intrinsic image decomposition algorithms for albedo estimation.

\paragraph{Evaluation of normal estimation.}

\begin{figure}[!h]
	\centering
	\includegraphics[width=0.48\textwidth]{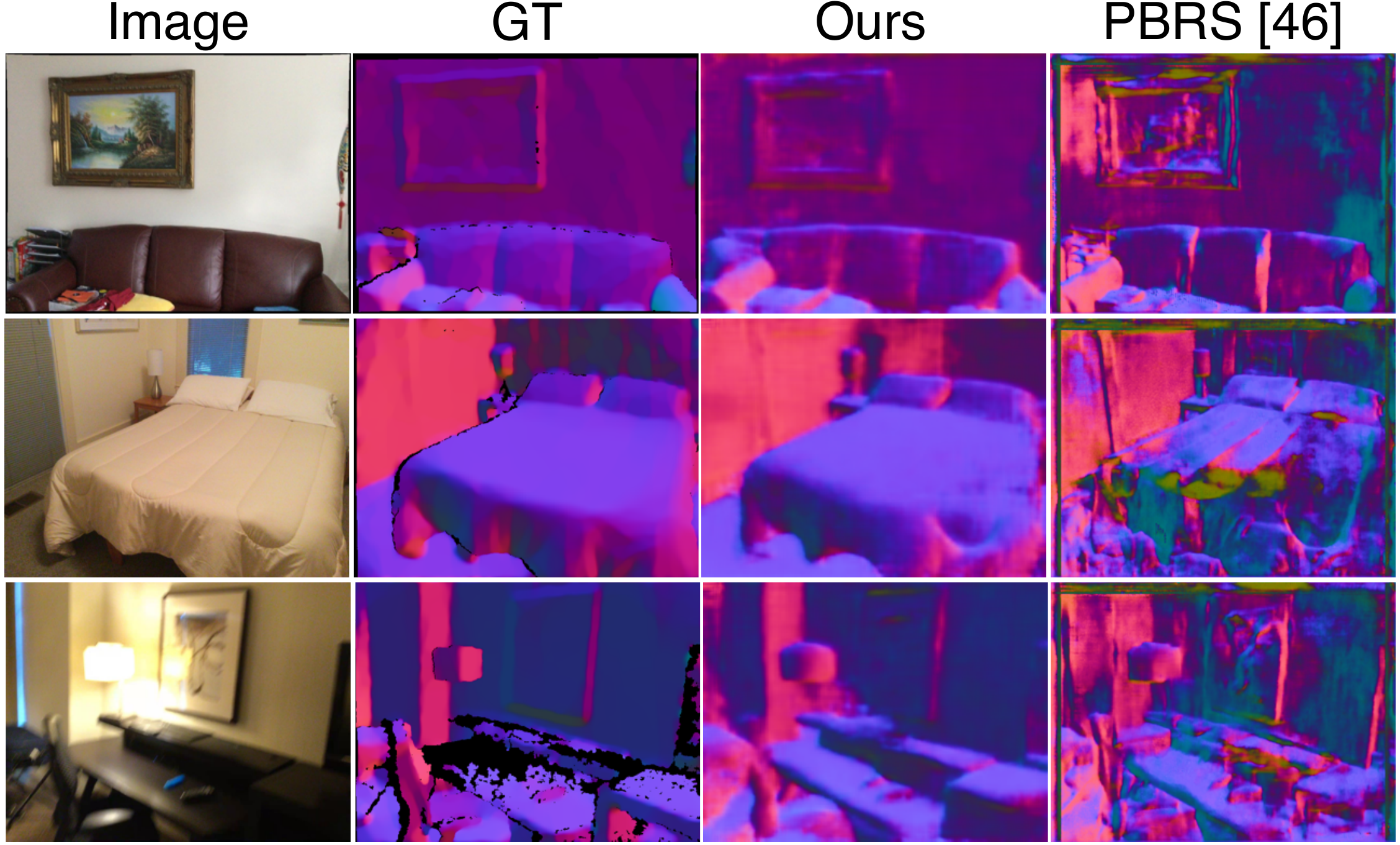}
	\caption{\small \textbf{Comparison with PBRS~\cite{zhang2016physically} on Scannet~\cite{dai2017scannet}.}}
	\label{fig:normal}
\end{figure}

We also compare with PBRS~\cite{zhang2016physically} which predicts only surface normals from an image. Both PBRS and `Ours' are trained on synthetic data and NYUv2~\cite{Silberman:ECCV12} (real data), and then tested on NYUv2, 7-scenes~\cite{shotton2013scene}, Scannet~\cite{dai2017scannet} and synthetic CG-PBR datasets. In Table~\ref{tab:nyu} we evaluate median angular error over the estimated albedo obtained by PBRS and our approach. Our method significantly outperforms PBRS on Scannet and CG-PBR, while slightly improving on 7-Scenes and doing slightly worse on NYUv2. This suggest that while PBRS overfits on NYUv2, our method shows significantly better generalization across datasets. This is because we are jointly reasoning about all components of the scene. Qualitative comparisons of normals predicted by our method and that of PBRS on Scannet dataset are shown in Fig.~\ref{fig:normal}).

\begin{table}[!h]
	\centering
	\captionsetup{justification=centering}
	\caption{\small \textbf{Median angular errors for surface normals.\\} (trained on synthetic and NYUv2)}
	\begin{tabular}{ccccc}
		\toprule
		& \small{NYUv2}        & \small{7-scenes} & \small{Scannet} & \small{CG-PBR}  \\ 
		\midrule
		PBRS~\cite{zhang2016physically} & \textbf{15.33\degree} & 25.65\degree & 30.39\degree & 27.84\degree \\
		\textbf{Ours}                           & 16.92\degree & \textbf{24.54\degree} & \textbf{21.10\degree} & \textbf{18.67\degree}\\ 
		\bottomrule
	\end{tabular}
	\label{tab:nyu}
\end{table}

\begin{table}[!h]
	\centering
	\caption{\small \textbf{Environment map estimation -- synth.~data.} (MAD)}
	\vspace{-.5em}
	\begin{tabular}{cccc}
		\toprule
		& Gardner ~\cite{gardner2017learning} & Ours & GT   \\ 
		\midrule
		\small{Env.\ map error} & 0.3162 & \textbf{0.1639} & -\\
		\small{Image recon.\ error} & 0.1640 & \textbf{0.1158} & 0.0949 \\
		\bottomrule
	\end{tabular}
	\label{tab:light_syn}
\end{table}

\begin{table}[!h]
	\centering
	\caption{\small \textbf{Environment map estimation -- real data.}}
	\vspace{-.5em}
	\begin{tabular}{ccc}
		\toprule
		& RMSE       & MAD   \\ 
		\midrule
		Gardner ~\cite{gardner2017learning} & 0.3109 & 0.2554  \\
		\textbf{Ours}                           & \textbf{0.2105}  & \textbf{0.1772} \\ 
		\bottomrule
	\end{tabular}
	\vspace{-.5em}
	\label{tab:light_real}
\end{table}

\paragraph{Evaluation of lighting estimation.} 

\begin{figure}[!h]
	\centering
	\includegraphics[width=0.45\textwidth]{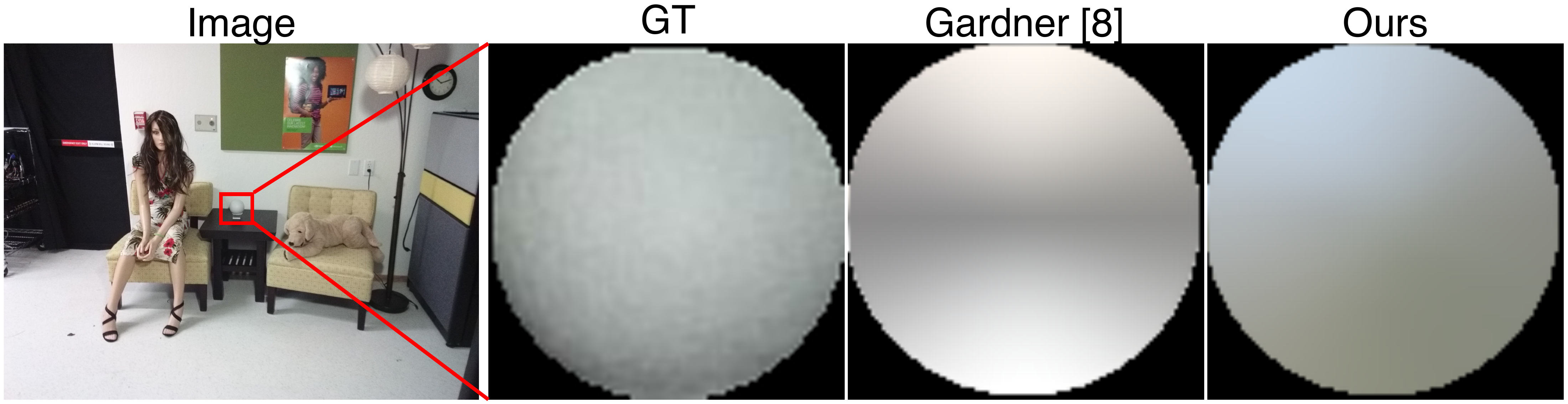}
	\caption{\small \textbf{Comparison with Gardner \etal~\cite{gardner2017learning}.} We collect 20 images of scenes with a diffuse ball, which is cropped as `GT'. Rendered diffuse ball with environment estimated by `Ours' outperforms  Gardner \etal~\cite{gardner2017learning} significantly.}
	\label{fig:lighting_real}
\end{figure}

We estimate an environment map of low spatial resolution from an image. We compare our estimated environment map with Gardner \etal~\cite{gardner2017learning}, the only existing method which also estimates a HDR environment map from a single indoor image using CNNs. We present a quantitative evaluation of the estimated environment map on synthetic data in Table \ref{tab:light_syn} (see Section \ref{sec:app_res}  for qualitative comparisons). For `Env. map error' we present the Mean Absolute Deviation (MAD) error w.r.t.\ `GT' (obtained by using $h_e(\cdot;\Theta_e)$, see Sec.~\ref{subsec:approach_synthetic}) weighted by solid angle, following ~\cite{weber2018learning}. We also compute MAD `Image recon.\ error' by using direct renderer $f_d(\cdot)$ (`GT' has a non-zero image reconstruction error, as it only captures distant-direct illumination.). For evaluating on real data, we collect 20 images of 4 scenes with varying illumination conditions with (also without) a diffuse ball inside the image, which serves as GT, as shown in Fig.~\ref{fig:lighting_real}. We estimate the environment map from the image taken without the diffuse ball using our method and that of Gardner \etal~\cite{gardner2017learning}. Then we render the diffuse ball with the estimated environment map and evaluate RMSE and MAD reconstruction error with the GT ball in Table~\ref{tab:light_real}. Our method significantly outperforms Gardner \etal~\cite{gardner2017learning} in estimating an environment map from a single image.

\section{Ablation Study}
\label{sec:result_ablation}

\paragraph{Role of RAR in self-supervised training.} 

The goal of RAR is to enable self-supervision on real images by capturing complex appearance effects that cannot be modeled by a direct renderer. RAR, along with the direct renderer, can reconstruct the image from its components, so that it can be used to train with a reconstruction loss. To show the effectiveness of RAR, we train IRN with (a) pseudo-supervision over albedo, normal and lighting (to handle ambiguity of decomposition as proposed in~\cite{sfsnetSengupta18}), and (b) weak supervision over normals or albedo, whenever available. Specifically, we train IRN: (i) on IIW with weak-supervision over albedo, with and without RAR; (ii) on NYUv2 with weak-supervision over normals, with and without RAR. Models trained on IIW with supervision over albedo are expected to produce better albedo estimates; models trained on NYUv2 with supervision over normals are expected to produce better normal estimates. We test these models on the CG-PBR and IIW to evaluate albedo and on NYUv2 and Scannet to evaluate normals. We report MAD error for the CG-PBR, the WHDR measure for the real IIW dataset and median angular error for the NYUv2 and Scannet in Table~\ref{tab:rar_norar}. Overall, RAR significantly improves the albedo and normal estimates across different datasets.

\begin{table}[!h]
	\centering
	\caption{\small \textbf{Role of RAR.} We evaluate albedo and normal estimation error by training IRN with (`Ours') and without (`w/o RAR') RAR using weak-supervisions over albedo (IIW) and normal (NYUv2).}
	\vspace{-.5em}
	\begin{tabular}{ccc|cc}
		\hline
		& \multicolumn{2}{c}{\small{Albedo}}     & \multicolumn{2}{c}{\small{Normal}}   \\ \hline
		& \small{CG-PBR} & \small{IIW}     & \small{NYUv2}         & \small{Scannet}       \\ \hline
		\multicolumn{1}{l|}{\small{IIW - Ours}}  & \textbf{0.130}  & \textbf{16.7\%} & \textbf{23.2\degree} & \textbf{28.4\degree} \\ 
		\multicolumn{1}{r|}{\small{ w/o RAR} }   & 0.265           & 21.6\%          & 60.2\degree          & 91.5\degree          \\ \hline
		\multicolumn{1}{l|}{\small{NYUv2 - Ours}}           & \textbf{0.134}  & \textbf{38.5\%} & \textbf{16.9\degree} & \textbf{21.1\degree} \\ 
		\multicolumn{1}{r|}{\small{w/o RAR} }      & 0.191           & 40.6\%          & 21.1\degree          & 87.1\degree          \\ \hline
	\end{tabular}
	\label{tab:rar_norar}
\end{table}

\begin{figure}[!h]
	\centering
	\includegraphics[width=0.48\textwidth]{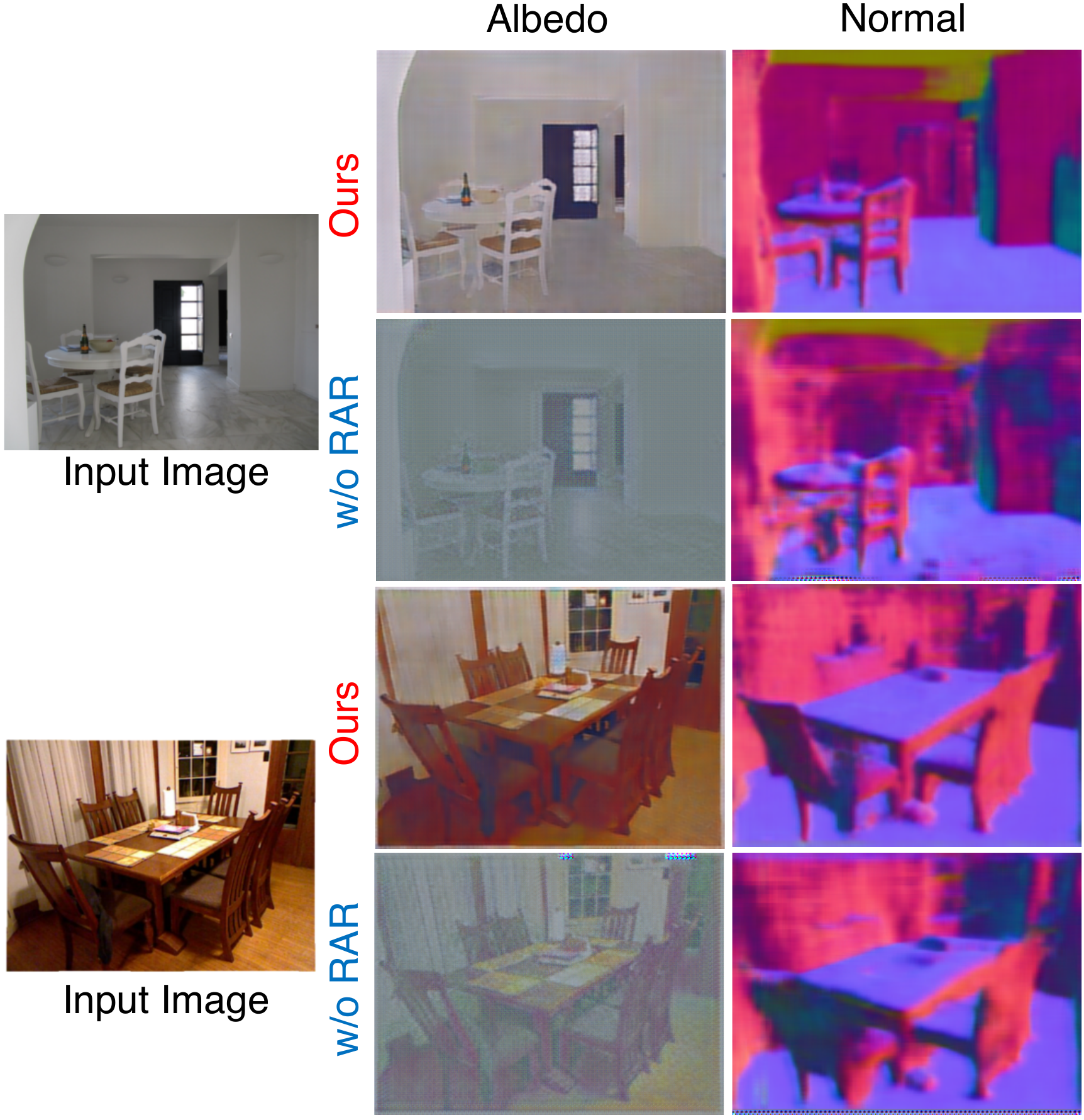}\vspace{-.75em}
	\caption{\small \textbf{Role of RAR in self-supervised training.}  We train IRN with (`Ours') and without (`w/o RAR') RAR on real data with weak-supervision over normals and albedo. Using RAR significantly improves the quality of estimated albedo and normal.}
	\label{fig:rar_norar}
\end{figure}

We further illustrate the importance of RAR with qualitative evaluations in Figure \ref{fig:rar_norar}. We train IRN with (`Ours') and without RAR (`w/o RAR'), with weak supervision over either albedo (WHDR loss on IIW for predicting albedo) or normal (L1 loss over normals in NYUv2 for predicting normals). Networks trained with supervision over one component, \eg normals, are always expected to produce better estimates of that component (normals) than the other (albedo). The normals are significantly improved when we use RAR. As for the albedo, using relative reflectance judgments without RAR produces very low contrast albedo. In the absence of RAR, the reconstruction loss used for self-supervised training cannot capture complex appearance effects, and thus it produces worse estimates of scene attributes.

\paragraph{Role of weak supervision.} 

\begin{figure}[!h]
	\centering
	\includegraphics[width=0.48\textwidth]{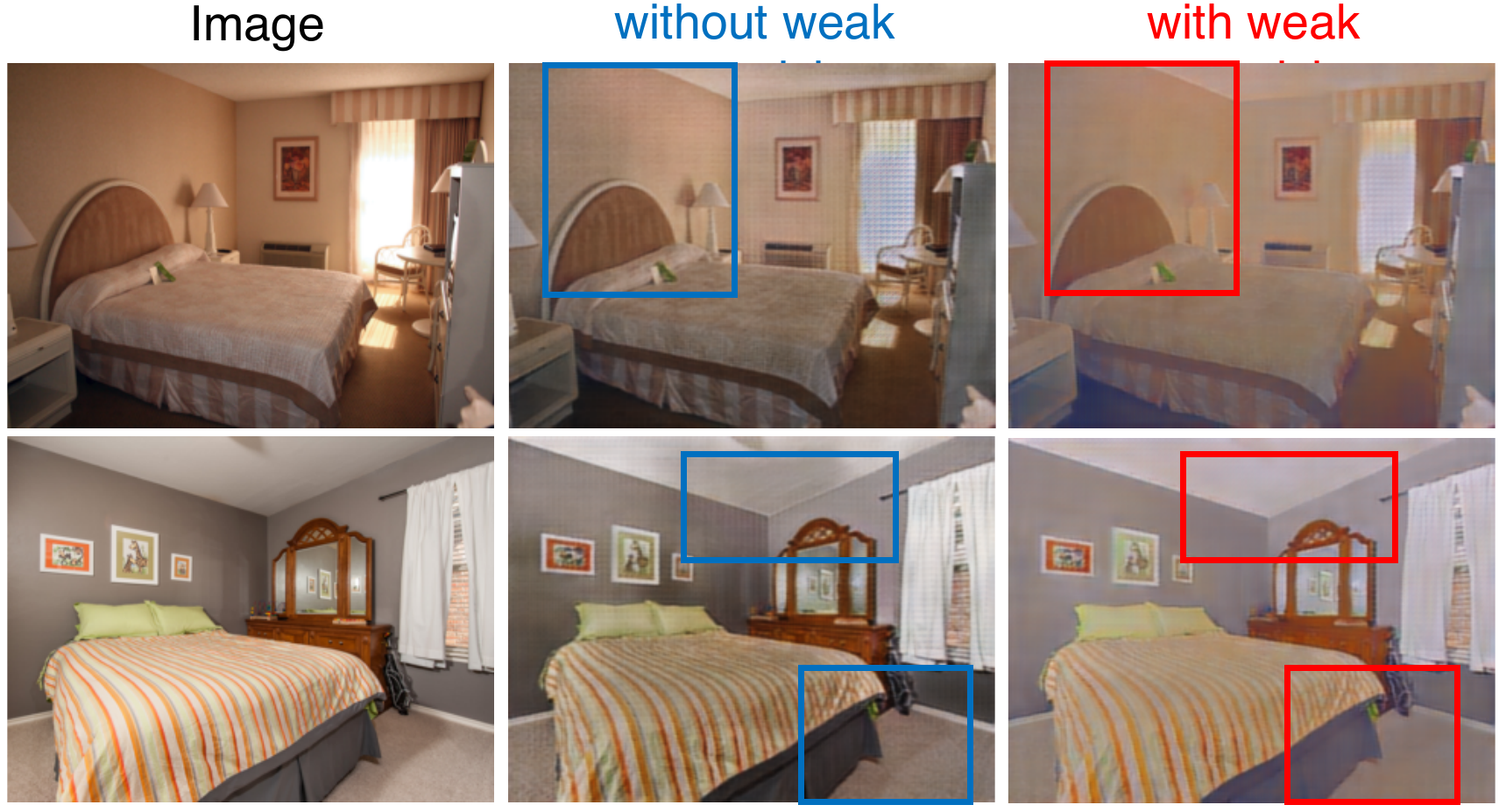}
	\caption{\small \textbf{Role of weak supervision.} We predict more consistent albedo across large objects like walls, floors and ceilings using pair-wise relative reflectance judgments from the IIW dataset~\cite{bell2014intrinsic}.}
	\label{fig:sparse}
\end{figure}

To evaluate the influence of weak supervision on inverse rendering, we train IRN with and without weak supervision over albedo (on IIW) and normals (NYUv2), respectively, as shown in Table~\ref{tab:weak_sup}. Weak supervision significantly reduces median angular error on the NYUv2 dataset and the WHDR metric on the IIW dataset. It also makes albedo prediction more consistent across large objects like walls, floors, and ceilings as shown in Fig.~\ref{fig:sparse}.

\begin{table}[!h]
	\centering
	\caption{\small \textbf{Role of weak supervision.} We train IRN with ('Ours') and without (`w/o RAR') RAR using weak-supervisions over albedo (IIW) and normal respectively (NYUv2). }
	\vspace{-.5em}
	\begin{tabular}{ccc}
		\toprule
		& Albedo (IIW)       & Normal (NYUv2) \\ 
		\midrule
		Ours & \textbf{16.7\%} & \textbf{16.9\degree} \\
		w/o weak sup.  & 32.7\% & 23.3\degree \\ 
		\bottomrule
	\end{tabular}
	\label{tab:weak_sup}
\end{table}

\section{Conclusion}
\label{sec:conclusion}

We present a learning-based approach for inverse rendering of an indoor scene from a single RGB image. Experimental results show our method outperforms prior works for estimating albedo, normal and lighting, which shows the effectiveness of joint learning. We propose a novel Residual Appearance Renderer (RAR) that can synthesize complex appearance effects such as inter-reflection, cast shadows, near-field illumination, and realistic shading. We show that this renderer is important for employing the self-supervised reconstruction loss to learn inverse rendering on real images. Although the goal of RAR in this paper is to enable self-supervision and improve inverse rendering estimations, we believe it is a good starting point for developing neural renderers that can handle object insertion. We also create a large-scale high-quality synthetic dataset CG-PBR with physically-based rendering, which we believe will be useful to the research community.

\textbf{Acknowledgment} Soumyadip Sengupta and David W. Jacobs are supported by the National Science Foundation under grant no. IIS-1526234.

\newpage

{\small
	\bibliographystyle{ieee}
	\bibliography{ref}

\begin{thebibliography}{10}\itemsep=-1pt

\bibitem{BarronTPAMI2015}
J.~T. Barron and J.~Malik.
\newblock Shape, illumination, and reflectance from shading.
\newblock {\em IEEE Transactions on Pattern Analysis and Machine Intelligence
  (TPAMI)}, 2015.

\bibitem{bell2014intrinsic}
S.~Bell, K.~Bala, and N.~Snavely.
\newblock Intrinsic images in the wild.
\newblock {\em ACM Transactions on Graphics (TOG)}, 33(4):159, 2014.

\bibitem{chaitanya2017interactive}
C.~R.~A. Chaitanya, A.~S. Kaplanyan, C.~Schied, M.~Salvi, A.~Lefohn,
  D.~Nowrouzezahrai, and T.~Aila.
\newblock Interactive reconstruction of monte carlo image sequences using a
  recurrent denoising autoencoder.
\newblock {\em ACM Transactions on Graphics (TOG)}, 36(4):98, 2017.

\bibitem{che2018inverse}
C.~Che, F.~Luan, S.~Zhao, K.~Bala, and I.~Gkioulekas.
\newblock Inverse transport networks.
\newblock {\em arXiv preprint arXiv:1809.10820}, 2018.

\bibitem{dai2017scannet}
A.~Dai, A.~X. Chang, M.~Savva, M.~Halber, T.~Funkhouser, and M.~Nie{\ss}ner.
\newblock Scannet: Richly-annotated 3d reconstructions of indoor scenes.
\newblock In {\em IEEE Conference on Computer Vision and Pattern Recognition
  (CVPR)}, pages 5828--5839, 2017.

\bibitem{eigen2015predicting}
D.~Eigen and R.~Fergus.
\newblock Predicting depth, surface normals and semantic labels with a common
  multi-scale convolutional architecture.
\newblock In {\em International Conference on Computer Vision (ICCV)}, pages
  2650--2658, 2015.

\bibitem{Fu2018DeepOR}
H.~Fu, M.~Gong, C.~Wang, K.~Batmanghelich, and D.~Tao.
\newblock Deep ordinal regression network for monocular depth estimation.
\newblock In {\em IEEE Conference on Computer Vision and Pattern Recognition
  (CVPR)}, 2018.

\bibitem{gardner2017learning}
M.-A. Gardner, K.~Sunkavalli, E.~Yumer, X.~Shen, E.~Gambaretto, C.~Gagn{\'e},
  and J.-F. Lalonde.
\newblock Learning to predict indoor illumination from a single image.
\newblock {\em ACM Transactions on Graphics (TOG)}, 36(6):176, 2017.

\bibitem{georgoulis2016delight}
S.~Georgoulis, K.~Rematas, T.~Ritschel, M.~Fritz, L.~Van~Gool, and
  T.~Tuytelaars.
\newblock {DeLight-Net}: Decomposing reflectance maps into specular materials
  and natural illumination.
\newblock {\em arXiv preprint arXiv:1603.08240}, 2016.

\bibitem{hold2017deep}
Y.~Hold-Geoffroy, K.~Sunkavalli, S.~Hadap, E.~Gambaretto, and J.-F. Lalonde.
\newblock Deep outdoor illumination estimation.
\newblock In {\em IEEE Conference on Computer Vision and Pattern Recognition
  (CVPR)}, volume~2, 2017.

\bibitem{Mitsuba}
W.~Jakob.
\newblock Mitsuba renderer, 2010.
\newblock http://www.mitsuba-renderer.org.

\bibitem{Karsch:SA:11}
K.~Karsch, V.~Hedau, D.~Forsyth, and D.~Hoiem.
\newblock Rendering synthetic objects into legacy photographs.
\newblock In {\em ACM Transactions on Graphics (SIGGRAPH Asia)}, pages
  157:1--157:12, 2011.

\bibitem{kato2018neural}
H.~Kato, Y.~Ushiku, and T.~Harada.
\newblock Neural {3D} mesh renderer.
\newblock In {\em IEEE Conference on Computer Vision and Pattern Recognition
  (CVPR)}, pages 3907--3916, 2018.

\bibitem{khan2006image}
E.~A. Khan, E.~Reinhard, R.~W. Fleming, and H.~H. B{\"u}lthoff.
\newblock Image-based material editing.
\newblock {\em ACM Transactions on Graphics (TOG)}, 25(3):654--663, 2006.

\bibitem{kim2017lightweight}
K.~Kim, J.~Gu, S.~Tyree, P.~Molchanov, M.~Nie{\ss}ner, and J.~Kautz.
\newblock A lightweight approach for on-the-fly reflectance estimation.
\newblock In {\em International Conference on Computer Vision (ICCV)}, pages
  20--28, 2017.

\bibitem{lafortune1994using}
E.~P. Lafortune and Y.~D. Willems.
\newblock Using the modified {Phong} reflectance model for physically based
  rendering.
\newblock 1994.

\bibitem{lettry2018darn}
L.~Lettry, K.~Vanhoey, and L.~Van~Gool.
\newblock {DARN}: a deep adversarial residual network for intrinsic image
  decomposition.
\newblock In {\em IEEE Workshop on Applications of Computer Vision (WACV)},
  2018.

\bibitem{Li:2018:DMC}
T.-M. Li, M.~Aittala, F.~Durand, and J.~Lehtinen.
\newblock Differentiable monte carlo ray tracing through edge sampling.
\newblock 37(6):222:1--222:11, 2018.

\bibitem{li2018cgintrinsics}
Z.~Li and N.~Snavely.
\newblock {CGIntrinsics}: Better intrinsic image decomposition through
  physically-based rendering.
\newblock {\em European Conference on Computer Vision (ECCV)}, 2018.

\bibitem{li2018learning}
Z.~Li and N.~Snavely.
\newblock Learning intrinsic image decomposition from watching the world.
\newblock In {\em IEEE Conference on Computer Vision and Pattern Recognition
  (CVPR)}, 2018.

\bibitem{li2018materials}
Z.~Li, K.~Sunkavalli, and M.~Chandraker.
\newblock Materials for masses: {SVBRDF} acquisition with a single mobile phone
  image.
\newblock In {\em European Conference on Computer Vision (ECCV)}, 2018.

\bibitem{li2018svreflectance}
Z.~Li, Z.~Xu, R.~Ramamoorthi, K.~Sunkavalli, and M.~Chandraker.
\newblock Learning to reconstruct shape and spatially-varying reflectance with
  a single image.
\newblock In {\em ACM Transactions on Graphics (SIGGRAPH Asia)}, 2018.

\bibitem{material2017}
G.~Liu, D.~Ceylan, E.~Yumer, J.~Yang, and J.-M. Lien.
\newblock Material editing using a physically based rendering network.
\newblock In {\em International Conference on Computer Vision (ICCV)}, 2017.

\bibitem{lombardi2016reflectance}
S.~Lombardi and K.~Nishino.
\newblock Reflectance and illumination recovery in the wild.
\newblock {\em IEEE Transactions on Pattern Analysis and Machine Intelligence
  (TPAMI)}, 38(1):129--141, 2016.

\bibitem{Meka:2018}
A.~Meka, M.~Maximov, M.~Zollhoefer, A.~Chatterjee, H.-P. Seidel, C.~Richardt,
  and C.~Theobalt.
\newblock {LIME}: Live intrinsic material estimation.
\newblock In {\em IEEE Conference on Computer Vision and Pattern Recognition
  (CVPR)}, June 2018.

\bibitem{narihira2015direct}
T.~Narihira, M.~Maire, and S.~X. Yu.
\newblock Direct intrinsics: Learning albedo-shading decomposition by
  convolutional regression.
\newblock In {\em International Conference on Computer Vision (ICCV)}, pages
  2992--2992, 2015.

\bibitem{Silberman:ECCV12}
P.~K. Nathan~Silberman, Derek~Hoiem and R.~Fergus.
\newblock Indoor segmentation and support inference from {RGBD} images.
\newblock In {\em European Conference on Computer Vision (ECCV)}, 2012.

\bibitem{nestmeyer2017reflectance}
T.~Nestmeyer and P.~V. Gehler.
\newblock Reflectance adaptive filtering improves intrinsic image estimation.
\newblock In {\em IEEE Conference on Computer Vision and Pattern Recognition
  (CVPR)}, page~4, 2017.

\bibitem{oxholm2012shape}
G.~Oxholm and K.~Nishino.
\newblock Shape and reflectance from natural illumination.
\newblock In {\em European Conference on Computer Vision (ECCV)}, pages
  528--541. Springer, 2012.

\bibitem{prados2006shape}
E.~Prados and O.~Faugeras.
\newblock Shape from shading.
\newblock In {\em Handbook of mathematical models in computer vision}, pages
  375--388. Springer, 2006.

\bibitem{ronneberger2015u}
O.~Ronneberger, P.~Fischer, and T.~Brox.
\newblock {U-Net}: Convolutional networks for biomedical image segmentation.
\newblock In {\em International Conference on Medical image computing and
  computer-assisted intervention (MICCAI)}, pages 234--241. Springer, 2015.

\bibitem{sfsnetSengupta18}
S.~Sengupta, A.~Kanazawa, C.~D. Castillo, and D.~W. Jacobs.
\newblock Sfsnet: Learning shape, refectance and illuminance of faces in the
  wild.
\newblock In {\em IEEE Conference on Computer Vision and Pattern Recognition
  (CVPR)}, 2018.

\bibitem{shelhamer2015scene}
E.~Shelhamer, J.~T. Barron, and T.~Darrell.
\newblock Scene intrinsics and depth from a single image.
\newblock In {\em International Conference on Computer Vision, Workshops
  (ICCV-W)}, pages 37--44, 2015.

\bibitem{shi2017learning}
J.~Shi, Y.~Dong, H.~Su, and X.~Y. Stella.
\newblock Learning non-lambertian object intrinsics across shapenet categories.
\newblock In {\em IEEE Conference on Computer Vision and Pattern Recognition
  (CVPR)}, pages 5844--5853, 2017.

\bibitem{shotton2013scene}
J.~Shotton, B.~Glocker, C.~Zach, S.~Izadi, A.~Criminisi, and A.~Fitzgibbon.
\newblock Scene coordinate regression forests for camera relocalization in
  rgb-d images.
\newblock In {\em IEEE Conference on Computer Vision and Pattern Recognition
  (CVPR)}, pages 2930--2937, 2013.

\bibitem{shu2017neural}
Z.~Shu, E.~Yumer, S.~Hadap, K.~Sunkavalli, E.~Shechtman, and D.~Samaras.
\newblock Neural face editing with intrinsic image disentangling.
\newblock In {\em IEEE Conference on Computer Vision and Pattern Recognition
  (CVPR)}, pages 5444--5453, 2017.

\bibitem{song2016ssc}
S.~Song, F.~Yu, A.~Zeng, A.~X. Chang, M.~Savva, and T.~Funkhouser.
\newblock Semantic scene completion from a single depth image.
\newblock {\em IEEE Conference on Computer Vision and Pattern Recognition
  (CVPR)}, 2017.

\bibitem{Taniai2018}
T.~Taniai and T.~Maehara.
\newblock Neural inverse rendering for general reflectance photometric stereo.
\newblock In {\em Intl. Conf. on Machine Learning (ICML)}, pages 20--28, 2017.

\bibitem{tappen2003recovering}
M.~F. Tappen, W.~T. Freeman, and E.~H. Adelson.
\newblock Recovering intrinsic images from a single image.
\newblock In {\em Advances in Neural Information Processing Systems (NIPS)},
  pages 1367--1374, 2003.

\bibitem{tunwattanapong2009interactive}
B.~Tunwattanapong and P.~Debevec.
\newblock Interactive image-based relighting with spatially-varying lights.
\newblock In {\em ACM Transactions on Graphics (SIGGRAPH)}, 2009.

\bibitem{wang2018joint}
T.~Wang, T.~Ritschel, and N.~Mitra.
\newblock Joint material and illumination estimation from photo sets in the
  wild.
\newblock In {\em International Conference on 3D Vision (3DV)}, pages 22--31,
  2018.

\bibitem{weber2018learning}
H.~Weber, D.~Pr{\'e}vost, and J.-F. Lalonde.
\newblock Learning to estimate indoor lighting from 3d objects.
\newblock In {\em International Conference on 3D Vision (3DV)}.

\bibitem{xu2018deep}
Z.~Xu, K.~Sunkavalli, S.~Hadap, and R.~Ramamoorthi.
\newblock Deep image-based relighting from optimal sparse samples.
\newblock {\em ACM Transactions on Graphics (TOG)}, 37(4):126, 2018.

\bibitem{yu2019inverserendernet}
Y.~Yu and W.~A. Smith.
\newblock Inverserendernet: Learning single image inverse rendering.
\newblock In {\em IEEE Conference on Computer Vision and Pattern Recognition
  (CVPR)}, pages 3155--3164, 2019.

\bibitem{zhang2016emptying}
E.~Zhang, M.~F. Cohen, and B.~Curless.
\newblock Emptying, refurnishing, and relighting indoor spaces.
\newblock {\em ACM Transactions on Graphics (TOG)}, 35(6):174, 2016.

\bibitem{zhang2018discovering}
E.~Zhang, M.~F. Cohen, and B.~Curless.
\newblock Discovering point lights with intensity distance fields.
\newblock In {\em IEEE Conference on Computer Vision and Pattern Recognition
  (CVPR)}, pages 6635--6643, 2018.

\bibitem{zhang2016physically}
Y.~Zhang, S.~Song, E.~Yumer, M.~Savva, J.-Y. Lee, H.~Jin, and T.~Funkhouser.
\newblock Physically-based rendering for indoor scene understanding using
  convolutional neural networks.
\newblock {\em IEEE Conference on Computer Vision and Pattern Recognition
  (CVPR)}, 2017.

\bibitem{zhou2017label}
H.~Zhou, J.~Sun, Y.~Yacoob, and D.~W. Jacobs.
\newblock Label denoising adversarial network ({LDAN}) for inverse lighting of
  face images.
\newblock In {\em IEEE Conference on Computer Vision and Pattern Recognition
  (CVPR)}, 2017.

\bibitem{zhou2015learning}
T.~Zhou, P.~Krahenbuhl, and A.~A. Efros.
\newblock Learning data-driven reflectance priors for intrinsic image
  decomposition.
\newblock In {\em International Conference on Computer Vision (ICCV)}, pages
  3469--3477, 2015.

\bibitem{zhuo2015indoor}
W.~Zhuo, M.~Salzmann, X.~He, and M.~Liu.
\newblock Indoor scene structure analysis for single image depth estimation.
\newblock In {\em IEEE Conference on Computer Vision and Pattern Recognition
  (CVPR)}, pages 614--622, 2015.

\end{thebibliography}
}

\vspace{10mm}

	\section{Appendix}
	\label{sec:appen}
	
	In this appendix, we provide more details of our network architecture and the loss functions along with additional qualitative evaluations. Specifically, in Section \ref{sec:arch} we discuss the details of the IRN and RAR network architectures for reproducibility. Details of our training loss functions on real data are provided in Section \ref{sec:loss}. In Section \ref{sec:res} we present additional qualitative evaluations.
	
	\subsection{Network Architectures}
	\label{sec:arch}
	
			\subsubsection{IRN}
	Our proposed Inverse Rendering Network (IRN), shown again in Figure~\ref{fig:networks} for reference, is trained on real data using the Residual Appearance Renderer (RAR), which learns to capture the complex appearance effects (\textit{e.g.} inter-reflection, cast shadows, near-field illumination, and realistic shading). 
	In the following, we describe the implementation details of IRN and RAR.
	
\begin{figure}[]
			\centering
			\includegraphics[width=0.48\textwidth]{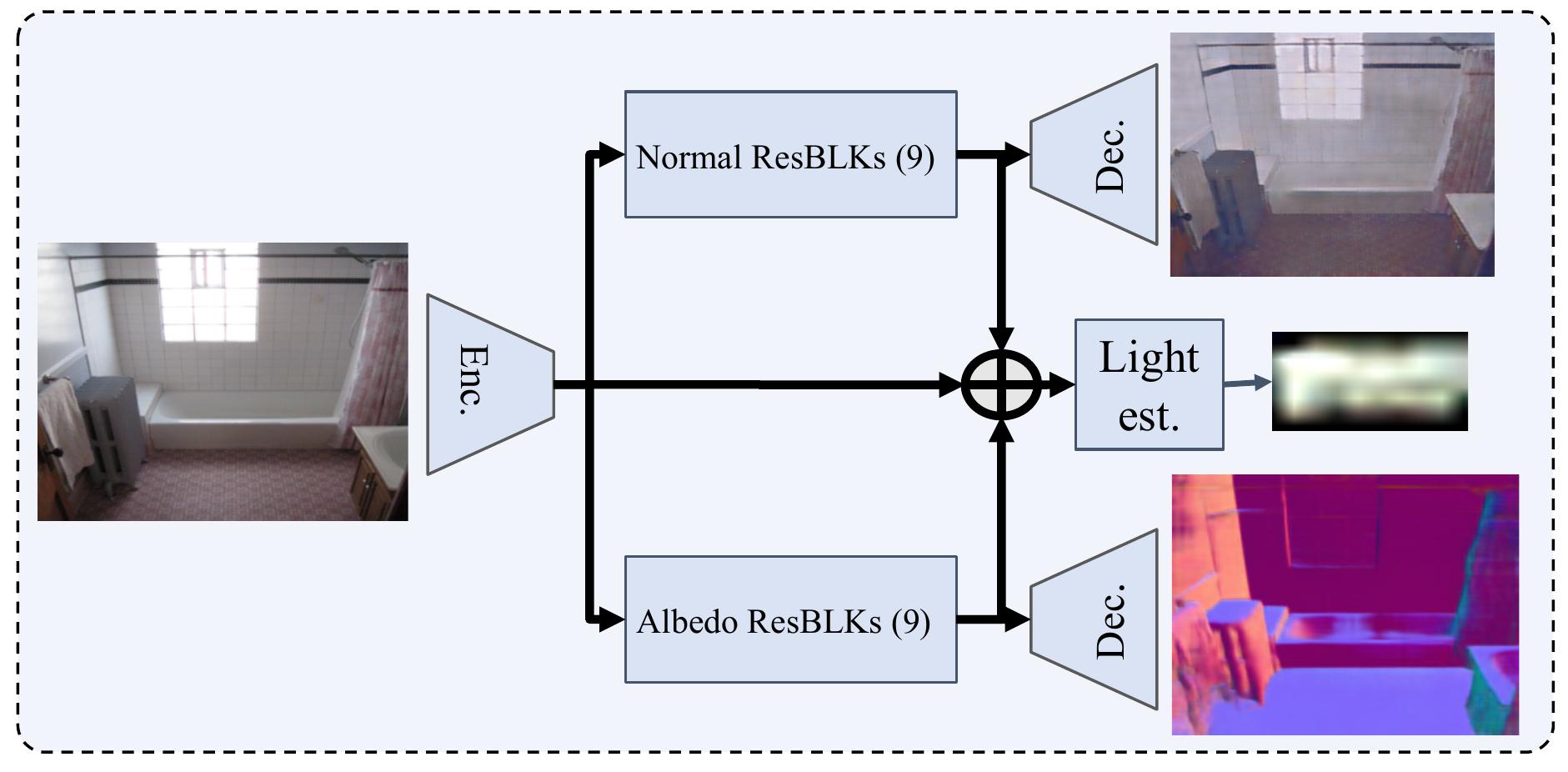}
			\caption{\small \textbf{IRN}.}
			\label{fig:net_irnD}	
		\end{figure}

	In Figure \ref{fig:net_irnD} we present the network architecture of IRN. The input to IRN is an image of spatial resolution $240\times320$, and the output is an albedo and normal map of same spatial resolution along with a $18\times36$ resolution environment map. Here we provide the details of the each block in IRN.
	
	\noindent
	\textbf{`Enc'}: C64(k7) - C*128(k3) - C*256(k3)\\
	`CN(kS)' denotes convolution layers with N $S \times S$ filters with stride 1, followed by Batch Normalization and ReLU. `C*N(kS)' denotes convolution layers with N $S \times S$ filters with stride 2, followed by Batch Normalization and ReLU. The output of `Enc' layer produces a blob of spatial resolution $256 \times 60 \times 80$.\\
	
	\noindent
	\textbf{`Normal ResBLKs'}: 9 ResBLK \\
	This consists of 9 Residual Blocks, `ResBLK's, which operate at a spatial resolution of $256 \times 60 \times 80$. Each `ResBLK' consists of Conv256(k3) - BN -ReLU - Conv256(k3) - BN, where `ConvN(kS)' and `BN' denote convolution layers with N $S \times S$ filters of stride 1 and Batch Normalization.
	
	
	\noindent
	\textbf{`Albedo ResBLKs'}: Same as `Normal Residual Blocks' (weights are not shared).\\
	
	\noindent
	\textbf{`Dec.'}: CD*128(k3)-CD*64(k3)-Co3(k7)\\
	`CD*N(kS)' denotes Transposed Convolution layers with N $S \times S$ filters with stride 2, followed by Batch Normalization and ReLU. `CN(kS)' denotes convolution layers with N $S \times S$ filters with stride 1, followed by Batch Normalization and ReLU. The last layer Co3k(7) consists of only convolution layers of 3 $7\times7$ filters, followed by Tanh layer. \\
	
	\noindent
	\textbf{`Light Est.'}: It first concatenates the responses of `Enc', `Normal ResBLKs' and `Albedo ResBLKs' to produce a blob of spatial resolution $768 \times 60 \times 80$. It is further processed by the following module:\\
	C256(k1) - C*256(k3) - C*128(k3) - C*3(k3) - BU(18,36)\\
	`CN(kS)' denotes convolution layers with N $S \times S$ filters with stride 1, followed by Batch Normalization and ReLU. `C*N(kS)' denotes convolution layers with N $S \times S$ filters with stride 2, followed by Batch Normalization and ReLU.  BU(18,36) upsamples the response to produce $18\times36\times3$ resolution environment map.

	\subsubsection{RAR}
	
	As shown in Figure \ref{fig:networks}, Residual Appearance Renderer (RAR) consists of a U-Net architecture and a convolution encoder. The U-Net consists of the following architecture, with normals and albedo as its input:\\
	`Encoder': C64(k3) - C*64(k3) - C*128(k3) - C*256(k3) - C*512(k3)\\
	`Decoder': CU512(k3) - CU256(k3) - CU128(k3) - CU64(k3) - Co3(k1)\\
	`CN(kS)' denotes convolution layers with N $S \times S$ filters with stride 1, followed by Batch Normalization and ReLU. `C*N(kS)' denotes convolution layers with N $S \times S$ filters with stride 2, followed by Batch Normalization and ReLU. `CUN(kS)' represents a bilinear up-sampling layer , followed by convolution layers with N $S \times S$ filters with stride 1, followed by Batch Normalization and ReLU. `Co3(k1)' consists of 3 $1 \times 1$ convolution filters to produce Normal or Albedo. Skip-connections exists between `C*N(k3)' layers of encoder and `CUN(k3)' layers of decoder. The encoder `Enc' that encodes image features to a latent $D=300$ dimensional subspace is given by:
	`Enc': C64(k7) - C*128(k3) - C*256(k3) - C128(k1) - C64(k3) - C*32(k3) - C*16(k3) - MLP(300)\\
	`CN(kS)' denotes convolution layers with N $S \times S$ filters with stride 1, followed by Batch Normalization and ReLU. `C*N(kS)' denotes convolution layers with N $S \times S$ filters with stride 2, followed by Batch Normalization and ReLU. MLP(300) takes the response of the previous layer and outputs a 300 dimensional feature, which is concatenated with the last layer of the U-Net `Encoder'.
	
	\subsubsection{Environment Map Estimator}
	
	As discussed in Section 3.1 of the main paper, the ground-truth environment map is estimated from the image, ground-truth albedo and normal using a deep network $h_e(\cdot,\Theta_e)$. The detailed architecture of this network is presented below:\\
	C64(k7) - C*128(k3) - C*256(k3) - 4 ResBLKS - C256(k1) - C*256(k3) - C*128(k3) - C*3(k3) - BU(18,36),\\
	where, `CN(kS)' denotes convolution layers with N $S \times S$ filters with stride 1, followed by Batch Normalization and ReLU. `C*N(kS)' denotes convolution layers with N $S \times S$ filters with stride 2, followed by Batch Normalization and ReLU. BU(18,36) upsamples the response to produce $18\times36\times3$ resolution environment map. Each `ResBLK' contains Conv256(k3) - BN -ReLU - Conv256(k3) - BN, where `ConvN(kS)' denotes convolution layers with N $S \times S$ filters of stride 1, `BN' denoted Batch Normalization.

	\subsection{Training Details}
	\label{sec:loss}
	
	\subsubsection{Training with weak supervision over albedo}
	
	IIW dataset presents relative reflectance judgments from humans. For any two points $R_1$ and $R_2$ on an image, a weighted confidence score classifies $R_1$ to be same, brighter or darker than $R_2$. We use these labels to construct a hinge loss for sparse supervision based on WHRD metric presented in \cite{bell2014intrinsic}. Specifically, if users predict $R_1$ to be darker than $R_2$ with confidence $w_t$, we use a loss $w_t \max (1+\delta - R_2/R_1,0)$. If $R_1$ and $R_2$ are predicted to have similar reflectance, we use $w_t [\max (R_1/R_2 -1 - \delta,0) + \max (R_2/R_1 -1 - \delta,0)]$. We observed empirically that this loss function performs better than WHRD metric, which is an L0 version of our loss. We train on real data with the following losses: (i) Psuedo-supervision loss over albedo ($L_a$), normal ($L_n$) and lighting ($L_e$) based on \cite{sfsnetSengupta18}, (ii) Photometric Reconstruction loss with RAR ($L_u$) (iii) Pair-wise weak supervision ($L_w$). Thus the net loss function is defined as:
	\begin{equation}
	L = 0.5*L_a + 0.5*L_n + 0.1*L_e + L_u + 30*L_w.
	\end{equation}
	
	\subsubsection{Training with weak supervision over normals}
	
	We also train on NYUv2 dataset with weak supervision over normals, obtained from Kinect depth data of the scene. We train with the following losses: (i) Psuedo-supervision loss over albedo ($L_a$) and lighting ($L_e$) based on \cite{sfsnetSengupta18}, (ii) Photometric Reconstruction loss with RAR ($L_u$) (iii) Supervision ($L_w$) over kinect normals. Thus the net loss function is defined as:
	\begin{equation}
	L = 0.2*L_a + 0.05*L_e + L_u + 20*L_w.
	\end{equation}

	\subsection{Our CG-PBR Dataset}
	
\begin{figure*}[t]
		\centering
		\includegraphics[width=1\textwidth]{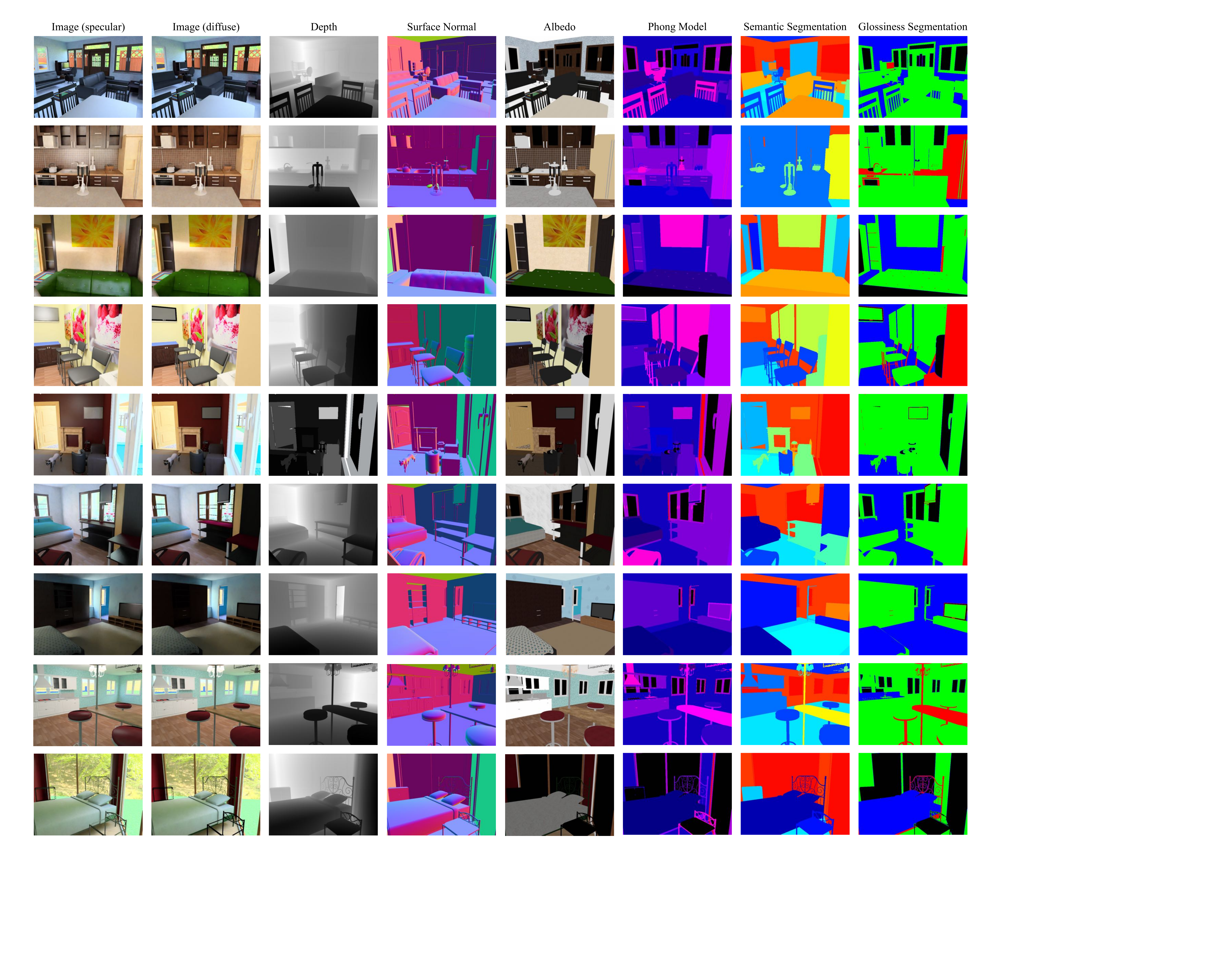}
		\caption{\small \textbf{Our CG-PBR Dataset.} We provide 235,893 images of a scene assuming specular and diffuse reflectance along with ground truth depth, surface normals, albedo, Phong model parameters, semantic segmentation and glossiness segmentation.}
		\vspace{5em}
		\label{fig:data_show1}	
	\end{figure*}

\begin{figure*}[t]
		\centering
		\includegraphics[width=1\textwidth]{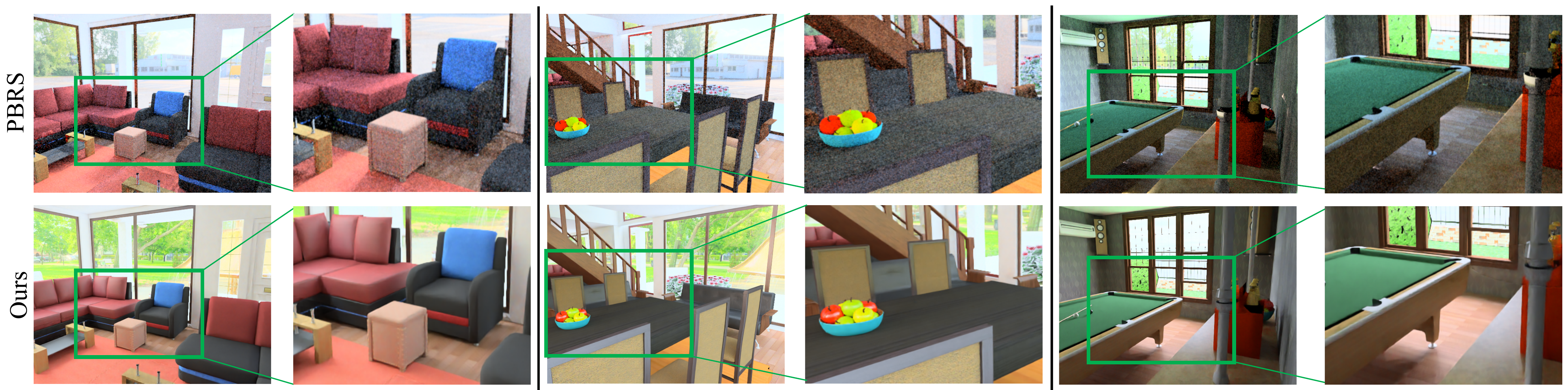}
		\caption{\small \textbf{Comparison with PBRS~\cite{zhang2016physically}.} Our dataset provides more photo-realistic and less noisy images with specular highlights under multiple lighting conditions.}
		\vspace{-1em}
		\label{fig:data_comp11}	
	\end{figure*}

\begin{figure*}[t]
		\centering
		\includegraphics[width=1\textwidth]{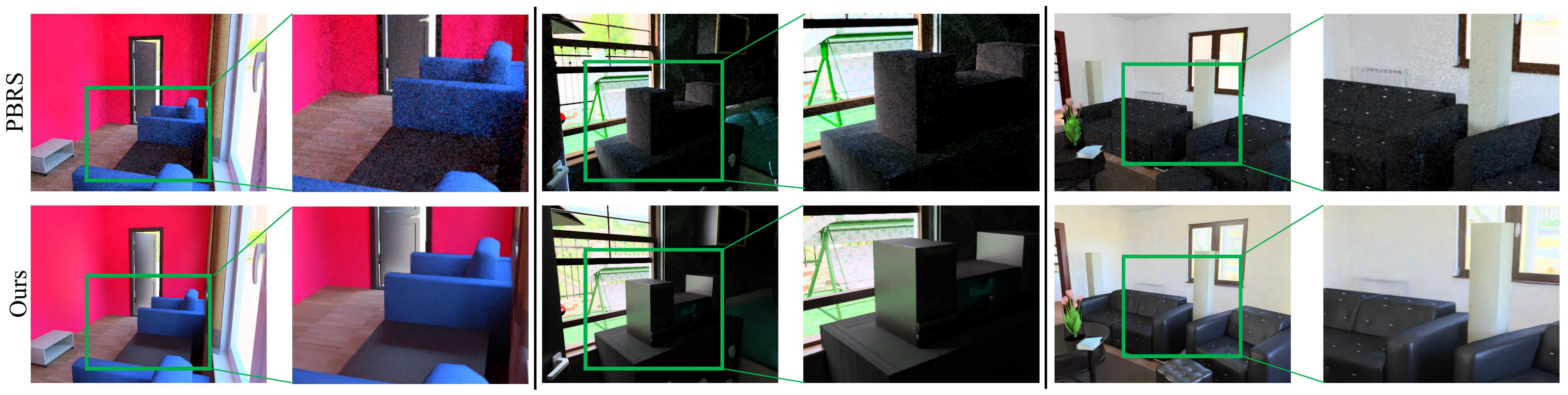}
		\caption{\small \textbf{Comparison with PBRS~\cite{zhang2016physically}.} Our dataset provides more photo-realistic and less noisy images with specular highlights under multiple lighting conditions.}
		\vspace{-1em}
		\label{fig:data_comp1}	
	\end{figure*}

	We present more example images of our CG-PBRS dataset in Figure \ref{fig:data_show1}. We also compare the renderings of our CG-PBR Dataset with that of PBRS~\cite{zhang2016physically}, under the same illumination condition in Figure \ref{fig:data_comp11} and \ref{fig:data_comp1}. CG-PBR provides more photo-realistic and less noisy images with specular highlights. Both CG-PBR and PBRS is rendered with Mitsuba~\cite{Mitsuba}. We will release the dataset upon publication.

	\subsection{More Experimental Results}
	\label{sec:app_res}
\begin{figure*}
		\centering
		\includegraphics[width=0.66\textwidth]{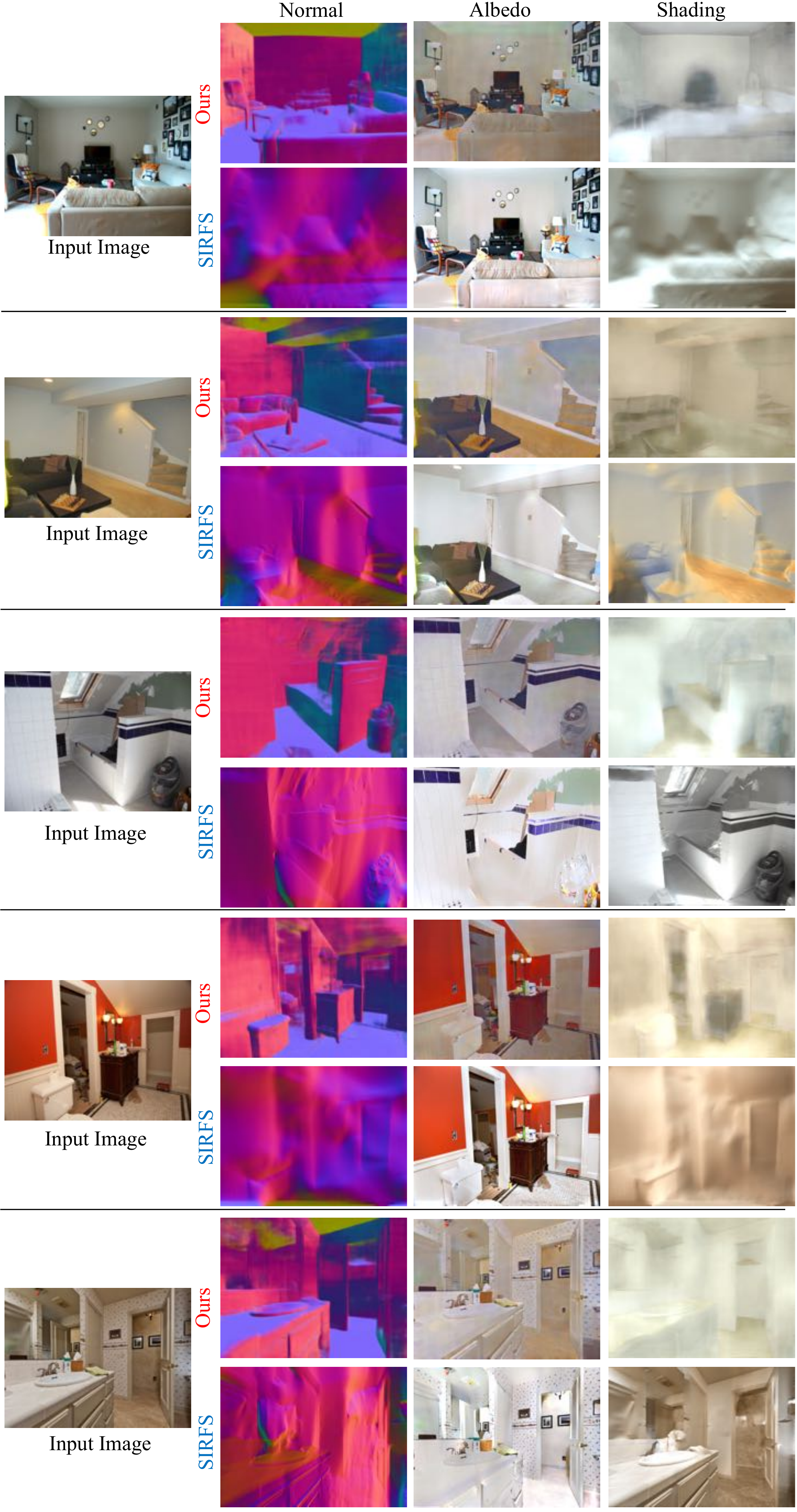}
		\caption{\small \textbf{Comparison with SIRFS~\cite{BarronTPAMI2015}.} Using deep CNNs our method performs better disambiguation of reflectance from shading and predicts better surface normals.}
		\vspace{-0.5em}
		\label{fig:sirfs1}
	\end{figure*}

	\paragraph{Comparison with SIRFS.} We present more detailed qualitative evaluations in this appendix. In Figure \ref{fig:sirfs1} we compare the results of our algorithm with that of SIRFS~\cite{BarronTPAMI2015}. SIRFS is an optimization-based method for inverse rendering, which estimates surface normals, albedo and spherical harmonics lighting from a single image. Compared to SIRFS we obtain more accurate normals
	and better disambiguation of reflectance from shading.

\begin{figure*}
		\centering
		\includegraphics[width=1\textwidth]{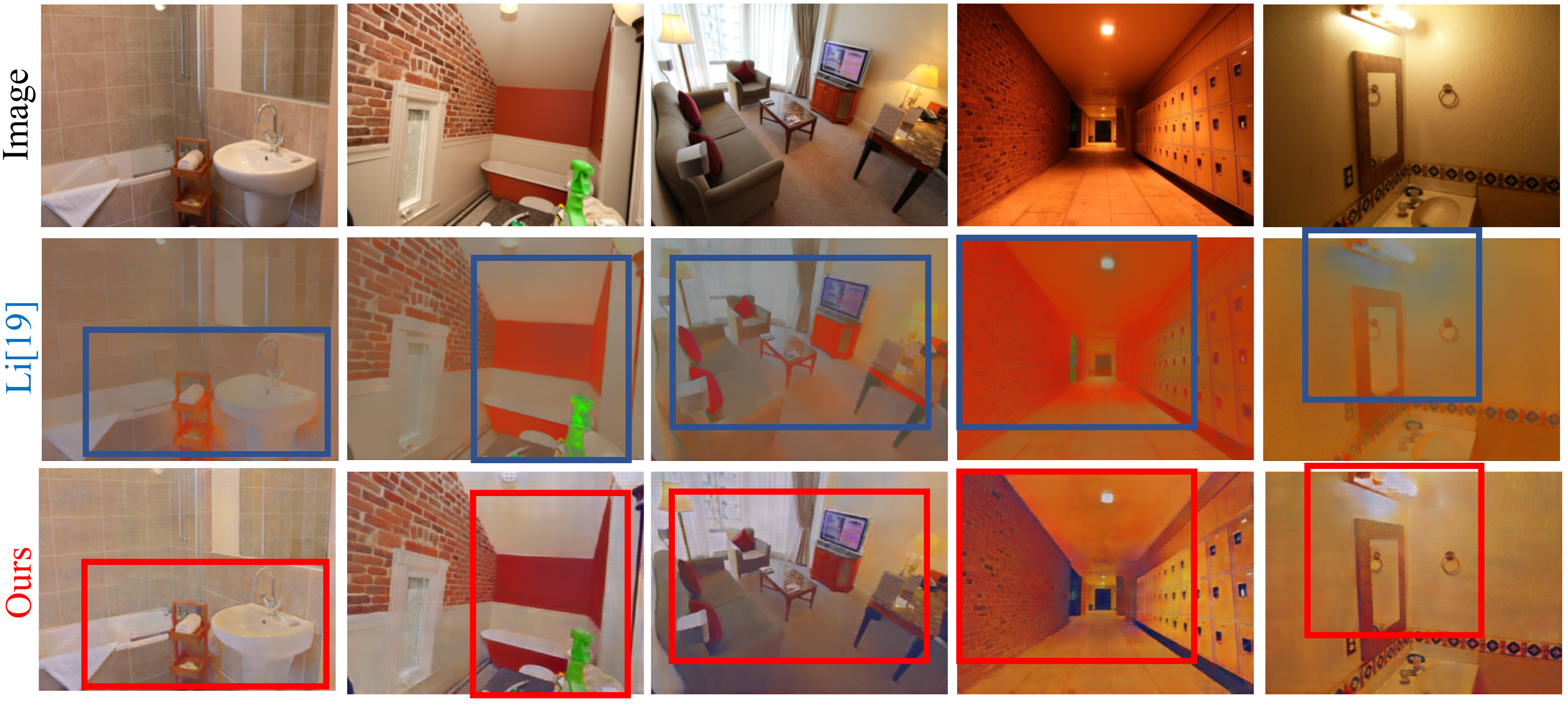}
		\caption{\small \textbf{Comparison with CGI (Li \textit{et. al.}~\cite{li2018cgintrinsics}) on IIW dataset. } In comparison with Li~\etal~\cite{li2018cgintrinsics}, our method performs better disambiguation of reflectance from shading and preserves the texture in the albedo.}
		\vspace{-1em}
		\label{fig:iiw_cgi1}
	\end{figure*}

\begin{figure*}
		\centering
		\includegraphics[width=1\textwidth]{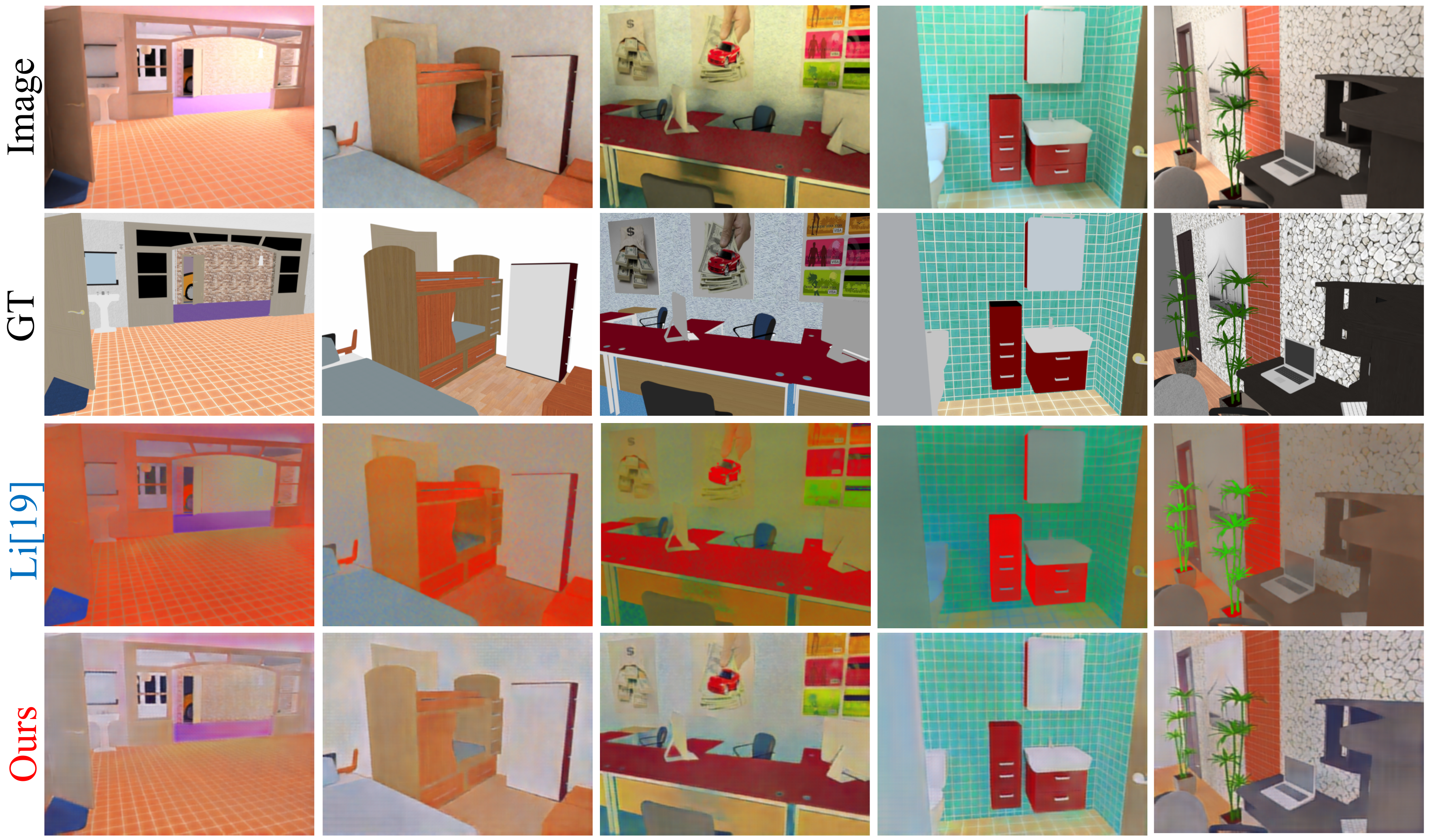}
		\caption{\small \textbf{Comparison with CGI (Li \textit{et. al.}~\cite{li2018cgintrinsics}) on CG-PBR (synthetic) dataset. } In comparison with Li~\etal~\cite{li2018cgintrinsics}, our method performs better disambiguation of reflectance from shading and preserves the texture in the albedo.}
		\vspace{-1em}
		\label{fig:syn_cgi}
	\end{figure*}

	\paragraph{Comparison with Li~\etal~\cite{li2018cgintrinsics}.} In Figure \ref{fig:iiw_cgi1} and \ref{fig:syn_cgi} we compare the albedo predicted by our method with that of Li~\etal~\cite{li2018cgintrinsics}, which performs intrinsic image decomposition of an image, on the real IIW and the synthetic CG-PBR dataset respectively. Intrinsic image decomposition methods do not explicitly recover geometry, illumination or glossiness of the material, but rather combine them together as shading. In contrast, our goal is to perform a complete inverse rendering which has a wider range of applications in AR/VR.

\begin{figure*}
		\centering
		\includegraphics[width=1\textwidth]{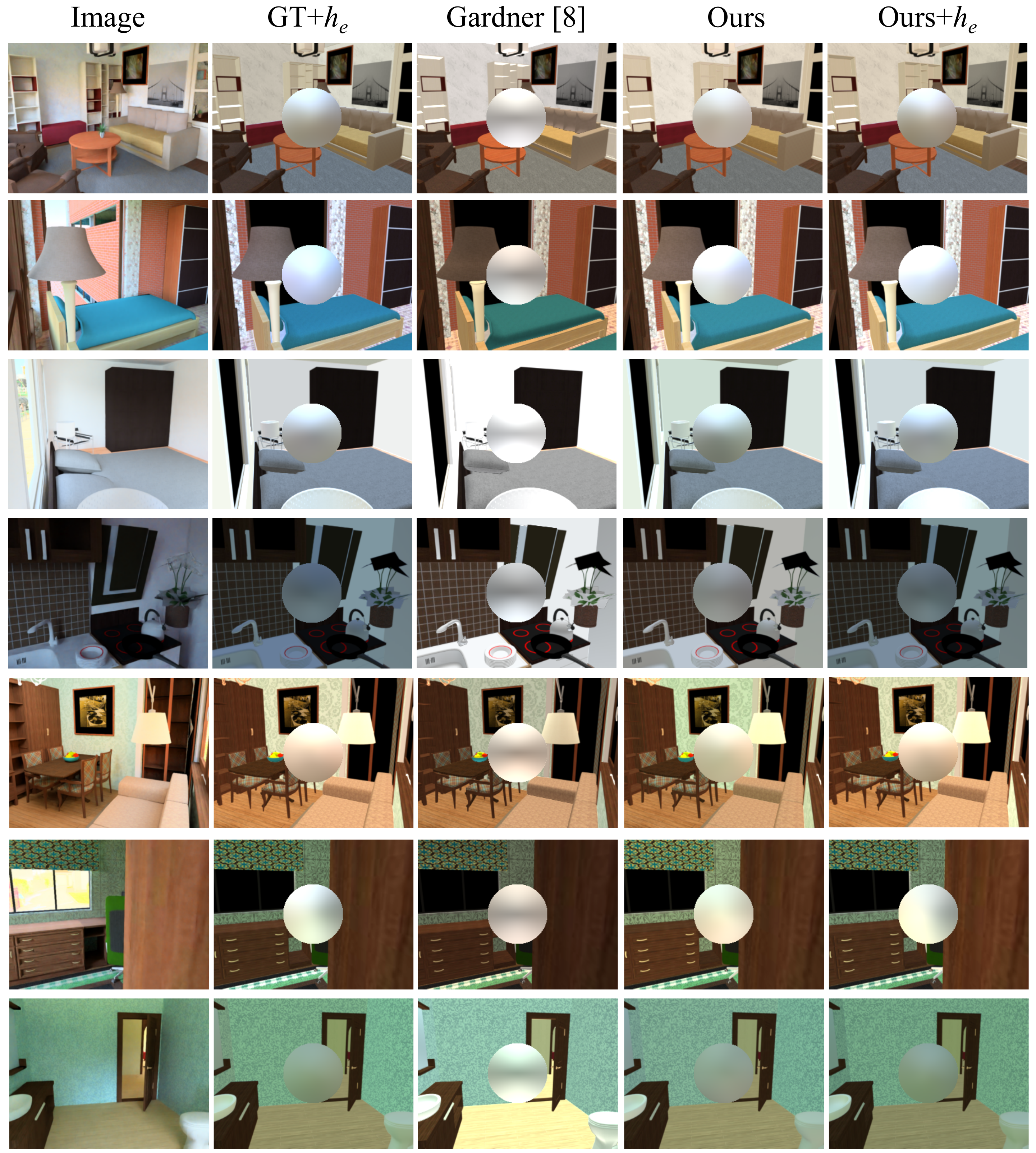}
		\caption{\small \textbf{Evaluation of lighting estimation.} We compare with Gardner \etal~\cite{gardner2017learning}. `GT+$h_e(\cdot)$' predicts lighting conditioned on the ground-truth normals and albedo. `Ours+$h_e(\cdot)$' predicts the environment map by conditioning it on the albedo and normals inferred by IRN.}
		\vspace{-1em}
		\label{fig:lighting}
	\end{figure*}

	\paragraph{Evaluation of lighting estimation.} In Figure \ref{fig:lighting} we present a qualitative evaluation of lighting estimation by inserting a diffuse hemisphere into the scene and rendering it with the inferred light from the image. We compare this with the method proposed by Gardner \etal~\cite{gardner2017learning}, which also estimates an environment map from a single indoor image. $h_e(\cdot,\Theta_e)$ is a deep network that predicts the environment map given the image, normals, and albedo. `GT+$h_e(\cdot)$' estimates the environment map given the image, ground-truth normals and albedo, and thus serves as an achievable upper-bound in the quality of the estimated lighting. `Ours' estimates environment map from an image with IRN. `Ours+$h_e(\cdot)$' predicts environment map by combining the inferred albedo and normals from IRN to predict lighting with $h_e(\cdot)$. Both `Ours' and `Ours+$h_e(\cdot)$' outperform Gardner \etal~\cite{gardner2017learning} as they seem to produce more realistic environment maps. `Ours+$h_e(\cdot)$' improves lighting estimation over `Ours' by utilizing the predicted albedo and normals to a greater degree. 
	
\begin{figure*}
		\centering
		\includegraphics[width=1\textwidth]{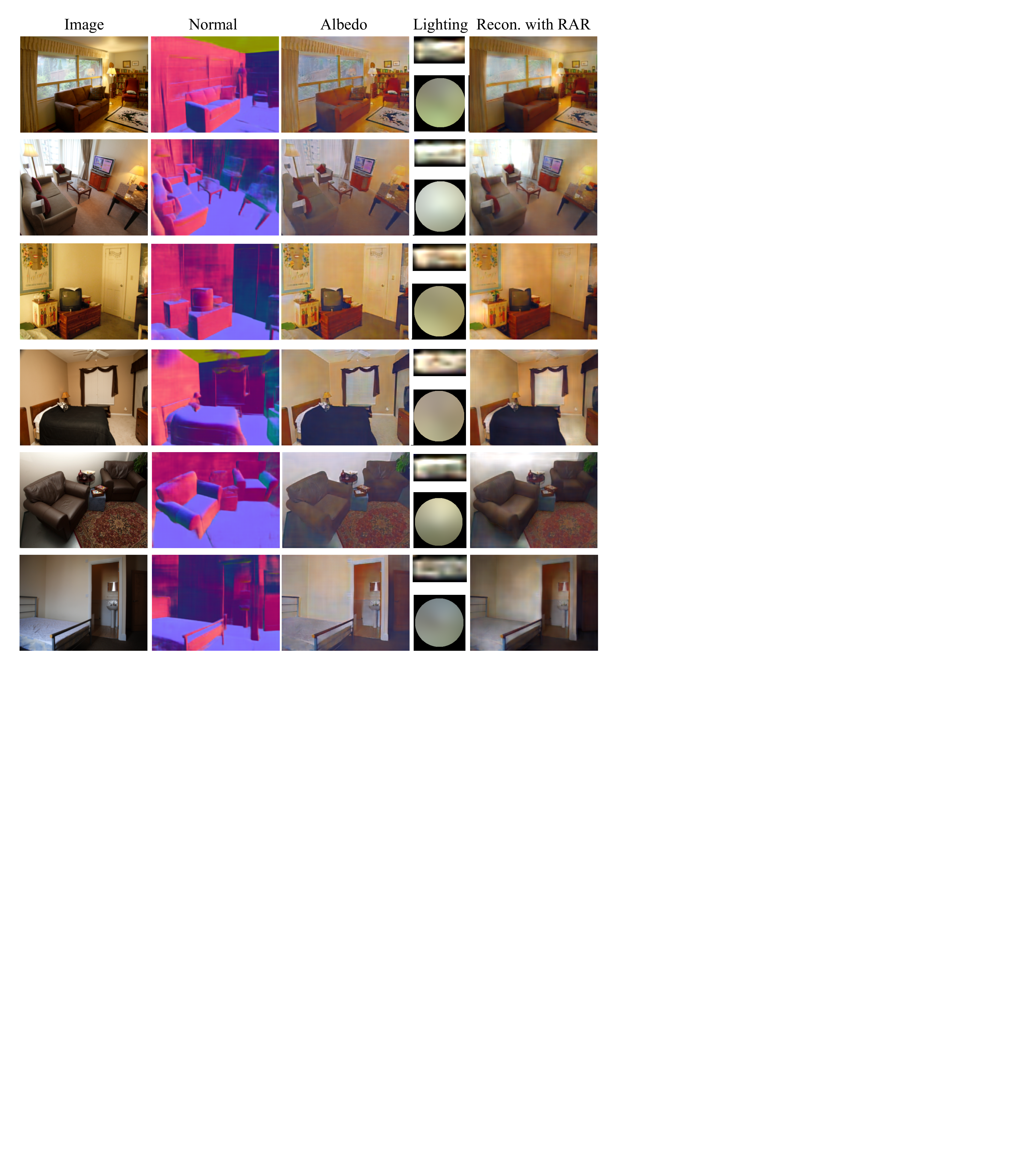}
		\caption{\small \textbf{Our Result.} We show the estimated intrinsic components; normals, albedo, and lighting predicted by the network, along with the reconstructed image with our direct renderer and the RAR.}
		\vspace{-1em}
		\label{fig:our_res}
	\end{figure*}

\begin{figure*}
		\centering
		\includegraphics[width=1\textwidth]{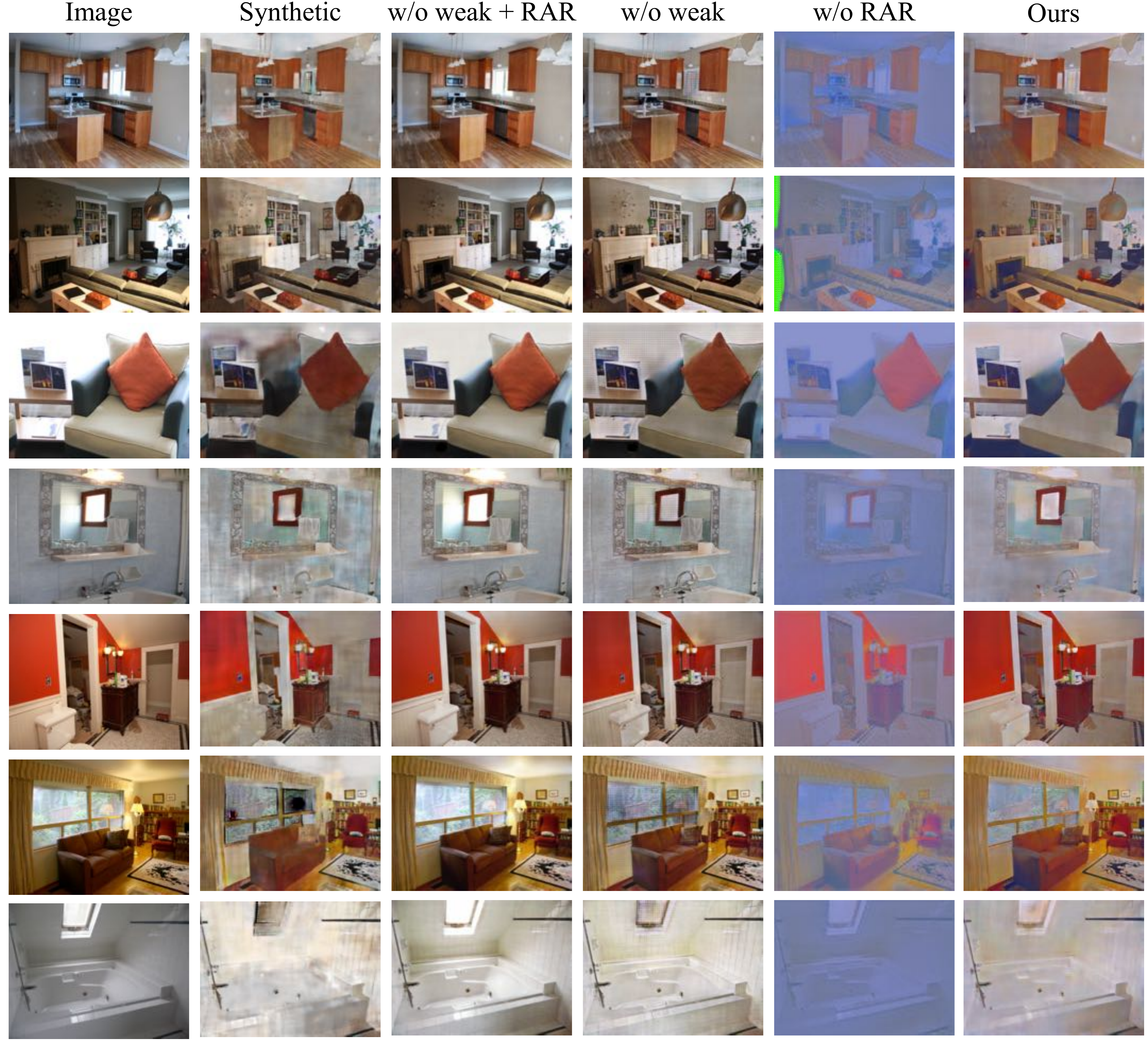}
		\caption{\small \textbf{Ablation Study}. We present the predicted albedo for each input image (in column 1) in column 2-6. We show the albedo predicted by IRN trained on our CG-PBR only in column 2. In column 6 (`Ours') we show the albedo predicted by IRN finetuned on real data with RAR and weak supervision over albedo. In column 4 and 5 we show the albedo predicted by IRN finetuned on real data without weak supervision (`w/o weak') and RAR (`w/o RAR') respectively. We present the albedo predicted by IRN finetuned without RAR and weak supervision on real data in column 3 (`w/o weak + RAR'). More images in Figure~\ref{fig:ab1}.}
		\vspace{-1em}
		\label{fig:ab}
	\end{figure*}

\begin{figure*}
		\centering
		\includegraphics[width=1\textwidth]{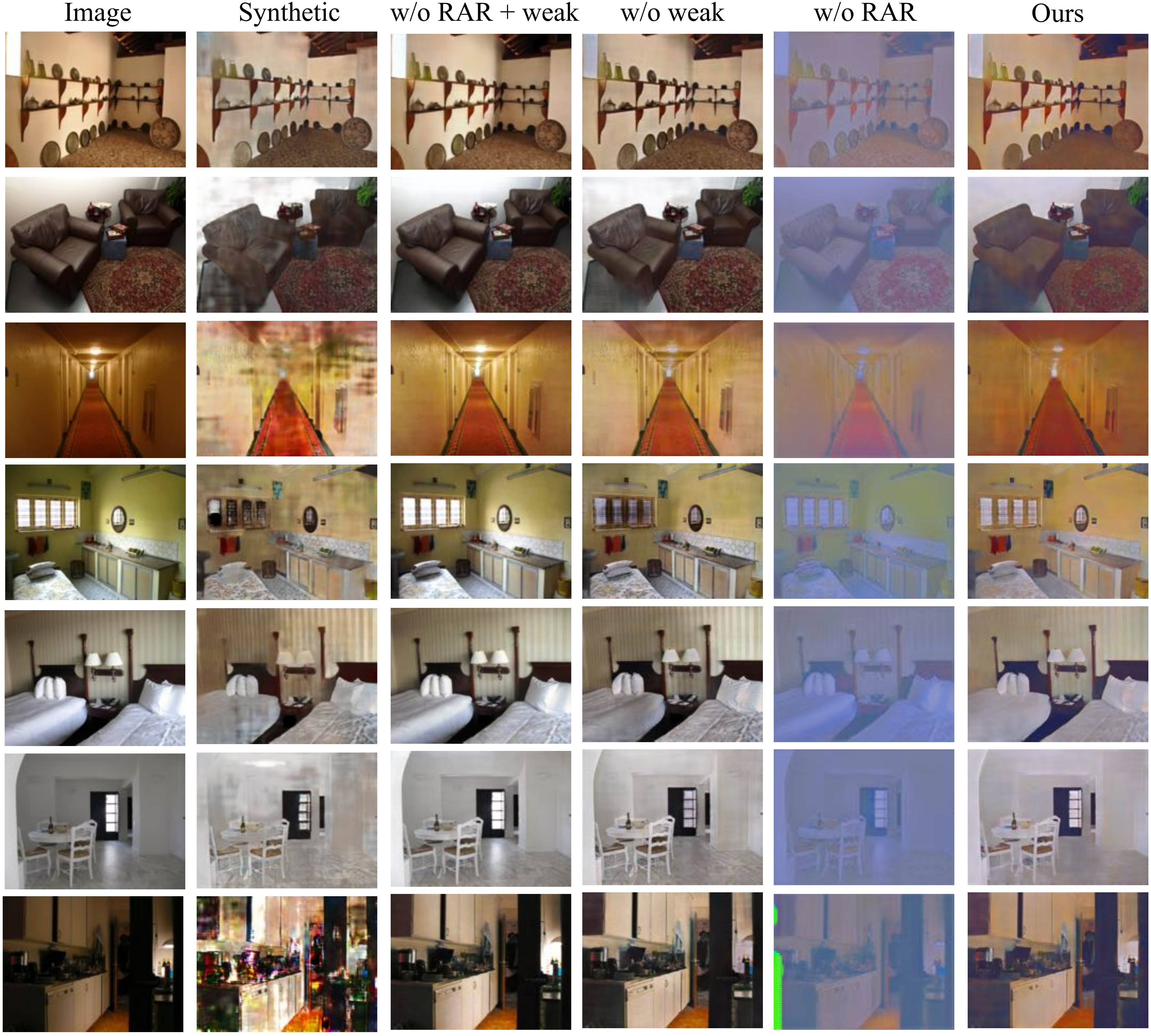}
		\caption{\small \textbf{Ablation Study}. We present the predicted albedo for each input image (in column 1) in column 2-6. We show the albedo predicted by IRN trained on our CG-PBR only in column 2. In column 6 (`Ours') we show the albedo predicted by IRN finetuned on real data with RAR and weak supervision over albedo. In column 4 and 5 we show the albedo predicted by IRN finetuned on real data without weak supervision (`w/o weak') and RAR (`w/o RAR') respectively. We present the albedo predicted by IRN finetuned without RAR and weak supervision on real data in column 3 (`w/o weak + RAR').}
		\vspace{-1em}
		\label{fig:ab1}
	\end{figure*}

	\paragraph{Our Results and Ablation study.} Figure~\ref{fig:our_res} shows examples of our results, with the albedo, normal and lighting predicted by the network, as well as the reconstructed image with the direct renderer and the proposed Residual Appearance Renderer (RAR). In Figure~\ref{fig:ab} and \ref{fig:ab1}, we perform a detailed ablation study of different components of our method. We show that it is important to train on real data, as networks trained on synthetic data fail to generalize well on real data. We also show that training our method without RAR (`w/o RAR') produces piece-wise smooth, low contrast albedo due to over-reliance on the weak supervision of pair-wise relative reflectance judgments. Training our network with only RAR, without any weak supervision (`w/o weak'), often fails to produce consistent albedo across large objects like walls, floor, ceilings etc. Finally, training without RAR and weak supervision (`w/o weak + RAR') produces albedo which contains the complex appearance effects like cast shadows, inter-reflections, highlights etc., as the reconstruction loss with direct renderer alone cannot model these effects.

\end{document}